\documentclass[letterpaper]{article} 
\usepackage[]{aaai24}  
\usepackage{times}  
\usepackage{helvet}  
\usepackage{courier}  
\usepackage[hyphens]{url}  
\usepackage{graphicx} 
\usepackage{natbib}  
\usepackage{caption} 
\usepackage{algorithm}
\usepackage{algorithmic}
\usepackage{chapterbib}
\usepackage{newfloat}
\usepackage{listings}
\DeclareCaptionStyle{ruled}{labelfont=normalfont,labelsep=colon,strut=off} 
\floatstyle{ruled}
\newfloat{listing}{tb}{lst}{}
\floatname{listing}{Listing}
\usepackage{cite,xspace}
\usepackage{amsmath,amssymb,amsfonts}
\usepackage{algorithmic}
\usepackage{algorithm}
\usepackage{float}
\usepackage{enumitem}
\usepackage{comment}
\usepackage{xspace}
\usepackage{booktabs, multirow, soul}
\usepackage{textcomp}
\usepackage[dvipsnames]{xcolor}
\usepackage[T1]{fontenc}
\usepackage[flushleft]{threeparttable}
\usepackage{subcaption}
\usepackage{mathtools}
\usepackage{url}
\makeatother

\usepackage{xr}
\usepackage{pdfpages}

\usepackage{placeins}

\title{Shape-conditioned 3D Molecule Generation via Equivariant Diffusion Models}
\author{
	Ziqi Chen\textsuperscript{\rm 1},
	Bo Peng\textsuperscript{\rm 1},
	Srinivasan Parthasarathy\textsuperscript{\rm 1,2},
	Xia Ning\textsuperscript{\rm 1,2,3}\thanks{Contact author},
}
\affiliations{
\textsuperscript{\rm 1}Computer Science and Engineering, The Ohio Sate University, Columbus, OH 43210\\
\textsuperscript{\rm 2}Translational Data Analytics Institute, The Ohio Sate University, Columbus, OH 43210\\
\textsuperscript{\rm 3}Biomedical Informatics, The Ohio Sate University, Columbus, OH 43210\\
chen.8484@buckeyemail.osu.edu, 
peng.307@buckeyemail.osu.edu, 
srini@cse.ohio-state.edu, 
ning.104@osu.edu
}

\newcommand{\etal}{\mbox{\emph{et al.}}\xspace}

\newcommand{\ziqi}[1]{\textcolor{blue}{#1}}

\newcommand{\method}{\mbox{$\mathop{\mathsf{ShapeMol}}\limits$}\xspace}
\newcommand{\methodenc}{\mbox{$\mathop{\mathsf{ShapeMol}\text{-}\mathsf{enc}}\limits$}\xspace}
\newcommand{\SE}{\mbox{$\mathop{\mathsf{SE}}\limits$}\xspace}
\newcommand{\SEE}{\mbox{$\mathop{\mathsf{SE}\text{-}\mathsf{enc}}\limits$}\xspace}
\newcommand{\SED}{\mbox{$\mathop{\mathsf{SE}\text{-}\mathsf{dec}}\limits$}\xspace}

\newcommand{\methoddiff}{\mbox{$\mathop{\mathsf{DIFF}}\limits$}\xspace}

\newcommand{\diffnoise}{\mbox{$\mathop{\mathsf{DIFF}\text{-}\mathsf{forward}}\limits$}\xspace}
\newcommand{\diffgenerative}{\mbox{$\mathop{\mathsf{DIFF}\text{-}\mathsf{backward}}\limits$}\xspace}
\newcommand{\methodwithguide}{\mbox{$\mathop{\mathsf{ShapeMol}\text{+}\mathsf{g}}\limits$}\xspace}

\newcommand{\eqgnn}{\mbox{$\mathop{\mathsf{EQ}\text{-}\mathsf{GNN}}\limits$}\xspace}
\newcommand{\invgnn}{\mbox{$\mathop{\mathsf{INV}\text{-}\mathsf{GNN}}\limits$}\xspace}

\newcommand{\squid}{\mbox{$\mathop{\mathsf{SQUID}}\limits$}\xspace}
\newcommand{\dataset}{\mbox{$\mathop{\mathsf{VS}}\limits$}\xspace}

\newcommand{\shape}{\mbox{$\mathop{\mathtt{s}}\limits$}\xspace}
\newcommand{\mol}{\mbox{$\mathop{\mathtt{M}}\limits$}\xspace}
\newcommand{\molx}{\mbox{$\mathop{\mathtt{M}_x}\limits$}\xspace}
\newcommand{\moly}{\mbox{$\mathop{\mathtt{M}_y}\limits$}\xspace}
\newcommand{\pc}{\mbox{$\mathop{\mathcal{P}}\limits$}\xspace}

\newcommand{\pos}{\mbox{$\mathop{\mathbf{x}}\limits$}\xspace}
\newcommand{\pointpos}{\mbox{$\mathop{\mathbf{z}}\limits$}\xspace}
\newcommand{\atomfeat}{\mbox{$\mathop{\mathbf{v}}$}\xspace}

\newcommand{\hiddenvec}{\mbox{$\mathop{\mathbf{h}}\limits$}\xspace}
\newcommand{\hiddenmat}{\mbox{$\mathop{\mathbf{H}}\limits$}\xspace}

\newcommand{\shapehiddenmat}{\mbox{$\mathop{\mathbf{H}^{\mathtt{s}}}\limits$}\xspace}

\newcommand{\cumalpha}{\mbox{$\mathop{\bar{\alpha}}\limits$}\xspace}

\newcommand{\alphav}{\mbox{$\mathop{\alpha^{\scriptsize{v}}}\limits$}\xspace}

\newcommand{\cfg}{\mbox{$\mathop{\mathsf{CFG}}\limits$}\xspace}

\newcommand{\shapesim}{\mbox{$\mathop{\mathsf{Sim}_{\mathtt{s}}}\limits$}\xspace}
\newcommand{\avgshapesim}{\mbox{$\mathop{\mathsf{avgSim}_{\mathtt{s}}}\limits$}}
\newcommand{\maxshapesim}{\mbox{$\mathop{\mathsf{maxSim}_{\mathtt{s}}}\limits$}}

\newcommand{\graphsim}{\mbox{$\mathop{\mathsf{Sim}_{\mathtt{g}}}\limits$}\xspace}
\newcommand{\avggraphsim}{\mbox{$\mathop{\mathsf{avgSim}_{\mathtt{g}}}\limits$}}
\newcommand{\maxgraphsim}{\mbox{$\mathop{\mathsf{maxSim}_{\mathtt{g}}}\limits$}}

\newcommand{\diversity}{\mbox{$\mathop{\mathsf{div}}\limits$}}

\newcommand{\qed}{\mbox{$\mathop{\mathsf{QED}}\limits$}\xspace}

\usepackage{bibentry}

\begin{document}

\maketitle

\begin{abstract}

Ligand-based drug design aims to identify novel drug candidates of similar shapes with known active molecules. 
In this paper, we formulated an \emph{in silico} shape-conditioned molecule generation problem to generate 3D molecule 
structures conditioned on the shape of a given molecule. 
To address this problem, we developed a
translation- and rotation-equivariant shape-guided generative model \method. 
%
%
%
\method consists of an equivariant shape encoder that maps molecular surface shapes into latent embeddings, and an equivariant
diffusion model that generates 3D molecules based on these embeddings.
Experimental results show that \method can generate novel, diverse, drug-like molecules that retain 3D molecular shapes similar to the given shape condition.
These results demonstrate the potential of \method in designing drug candidates of desired 3D shapes binding to
protein target pockets. 
\end{abstract}

\vspace{-10pt}
\section{Introduction}

Generating novel drug candidates is a critical step in drug discovery to identify possible 
therapeutic solutions.
Conventionally, this process is characterized based on knowledge and experience from medicinal chemists, 
and is resource- and time-consuming.
Recently, computational approaches to molecule generation have been developed
to accelerate the conventional paradigm.
Existing molecular generative models largely focus on generating either molecule SMILES strings or molecular graphs~\cite{GmezBombarelli2018, jin18jtvae,Chen2021modof}, with a recent 
shift towards 3D molecular structures.
Several models~\cite{luo2021sbdd, peng22pocket2mol,guan2023targetdiff} 
have been designed to generate 3D molecules 
conditioned on the protein targets,
aiming to facilitate structured-based drug design (SBDD)~{\cite{Batool_2019}}, 
given that molecules exist in 3D space and the efficacy of drug molecules depends on their 3D structures fitting into protein pockets. 
%
However, SBDD relies on the availability of high-quality 3D structures of protein binding pockets, 
which are lacking for many targets~\cite{Zheng_2013}.
%

Different from SBDD, ligand-based drug design (LBDD)~{\cite{Acharya2011}} 
utilizes ligands known to interact with a protein target, 
and does not require knowledge of protein structures.
In LBDD, shape-based virtual screening tools such as ROCS~\cite{Hawkins2006} have been 
widely used to identify molecules with similar shapes to known ligands 
by enumerating molecules in chemical libraries.
However, 
virtual screen tools cannot probe the novel chemical space.
Therefore, it is highly needed to develop generative methods to generate novel molecules with desired 
3D shapes. 


In this paper, we present a novel generative model for 3D molecule generation conditioned on given 3D shapes.
Our method, denoted to as \method, employs an equivariant shape embedding module to map 3D molecule surface shapes
into shape latent embeddings.
It then uses a conditional diffusion generative model to generate molecules conditioned on the shape latent 
embeddings, by iteratively denoising atom positions and atom features (e.g., atom type and aromaticity).
During molecule generation, \method can utilize additional shape guidance by pushing the predicted atoms far from 
the condition shapes to those shapes. \method with shape guidance is denoted as \methodwithguide.
The major contributions of this paper are as follows:
\begin{itemize}
\item To the best of our knowledge, 
\method is the first diffusion-based method for 3D molecule generation conditioned
on 3D molecule shapes.
\item \method leverages a new equivariant shape embedding module to 
learn 3D surface shape embeddings from cloud points sampled over molecule surfaces.
%
\item \method uses a novel conditional diffusion model to generate 3D molecule structures. The diffusion model
is equivariant to the translation and rotation of molecule shapes. A new weighting scheme over diffusion steps
is developed to ensure accurate molecule shape prediction. 
\item \method utilizes new shape guidance to direct the generated molecules to better fit the 
shape condition. 
%
%
\item \methodwithguide achieves the highest average 3D shape similarity between the generated molecules and condition molecules, compared to the state-of-the-art baseline. 
%
\end{itemize}

For reproducibility purposes, detailed parameters in all the experiments, code and data are reported in Supplementary Section~\ref{supp:experiments:parameters}.


\section{Related Work}

\subsection{Molecule Generation}

A variety of deep generative models have been developed to generate molecules using various molecule representations, 
incliuding generating SMILES string representations~\cite{GmezBombarelli2018}, or 2D molecular graph representations~\cite{jin18jtvae,Chen2021modof}.
Recent efforts have been dedicated to the generation of 3D molecules. 
%
These 3D molecule generative models can be divided into two categories: autoregressive models and non-autoregressive models.
Autoregressive models generate 3D molecules by sequentially adding atoms into the 3D space 
~\cite{luo2021sbdd, peng22pocket2mol}.
%
%
While these models ensure the validity and connectivity of generated molecules, any errors made in sequential predictions could accumulate in subsequent predictions.
Non-autoregressive models generate 3D molecules 
using flow-based methods~\cite{satorras2021} or diffusion methods~\cite{guan2023targetdiff}.
In these models, atoms are generated or adjusted all together. 
%
For example, Hoogeboom \etal~\shortcite{hoogeboom22diff} developed an equivariant diffusion model, 
in which an equivariant network is employed to
jointly predict both the positions and features of all atoms.
%

\subsection{{Shape-Conditioned Molecule Generation}}

Following the idea of ligand-based drug design (LBDD)~\cite{Acharya2011}, previous work has been focused on 
generating molecules with similar 3D shapes to those of efficacy,
based on the observation that structurally similar molecules tend to have similar properties.
Papadopoulos \etal~\shortcite{Papadopoulos2021} developed a reinforcement learning method 
to generate SMILES strings of molecules that are similar to known antagonists of DRD2 receptors 
in 3D shapes and pharmacophores.
%
Imrie \etal~\shortcite{Imrie2021} generated 2D molecular graphs conditioned on 3D pharmacophores using a graph-based autoencoder.
However, there is limited work that generates 3D molecule structures conditioned on 3D shapes.
Adams and Coley~\shortcite{adams2023equivariant} developed a shape-conditioned generative framework
\squid for 3D molecule generation.
{\squid} learns a variational autoencoder to generate fragments conditioned on given 3D shapes, and decodes molecules by sequentially attaching fragments with fixed bond lengths and angles.
%
%
%
%
While LBDD plays a vital role in drug discovery, the problem of generating 3D molecule structures
conditioned on 3D shapes is still under-addressed.

%
%


\section{Definitions and Notations}

\subsection{Problem Definition}

Following Adams and Coley \shortcite{adams2023equivariant}, we focus on the 3D molecule generation \textit{conditioned on} the shape of a given molecule (e.g., a ligand).
Specifically, we aim to generate a new molecule \moly, conditioned on the 3D shape of a given molecule \molx,  
such that 
1) {\moly} is similar to {\molx} in their 3D shapes, measured by $\shapesim(\shape_x, \shape_y)$, 
where \shape is the 3D shape of \mol; 
and 
2) {\moly} is dissimilar to {\molx} in their 2D molecular graph structures, measured by $\graphsim(\molx, \moly)$. 
This conditional 3D shape generation problem is motivated by the fact that 
in ligand-based drug design, it is desired to find chemically diverse and novel molecules that share similar shapes and similar activities with known active ligands~\cite{Ripphausen2011}.
%
Such chemically diverse and novel molecules could expand the search space for drug candidates and potentially enhance the 
development of effective drugs.
%

%

\subsection{Representations and Notations}


We represent a molecule \mol
as a set of atoms \mbox{$\mol = \{a_1, a_2, \cdots, a_{\scriptsize{|\mol|}}| a_i = (\pos_i, \atomfeat_i)\}$},
where $|\mol|$ is the number of atoms in \mol; $a_i$ is the $i$-th atom in \mol;  
$\pos_i\in \mathbb{R}^3$ represents the 3D coordinates of $a_i$; 
and $\atomfeat_i\in \mathbb{R}^{K}$ is $a_i$'s one-hot atom feature vector indicating the atom type and its
aromaticity.
Following Guan \etal~\shortcite{guan2023targetdiff}, bonds between atoms can be uniquely determined by the atom types and the atomic distances among atoms. 
We represent the 3D surface shape {\shape} of a molecule \mol as a point cloud constructed by sampling points over the molecular surface.
Details about the construction of point clouds from the surface of molecules are available in Supplementary Section~\ref{supp:point_clouds}.
We denote the point cloud as $\pc=\{z_1, z_2, \cdots\, z_{\scriptsize{|\pc|}} | z_j = (\pointpos_j)\}$, 
where $|\pc|$ is the number of points in $\pc$;
$z_j$ is the $j$-th point; and 
$\pointpos_j\in \mathbb{R}^3$ represents the 3D coordinates of $z_j$.
We denote the latent embedding of $\pc$ as $\shapehiddenmat\in \mathbb{R}^{d_p\times 3}$, where $d_p$ is the dimension of the latent embedding.

\section{Method}

\begin{figure}
\centering
\resizebox{\linewidth}{!}{\includegraphics{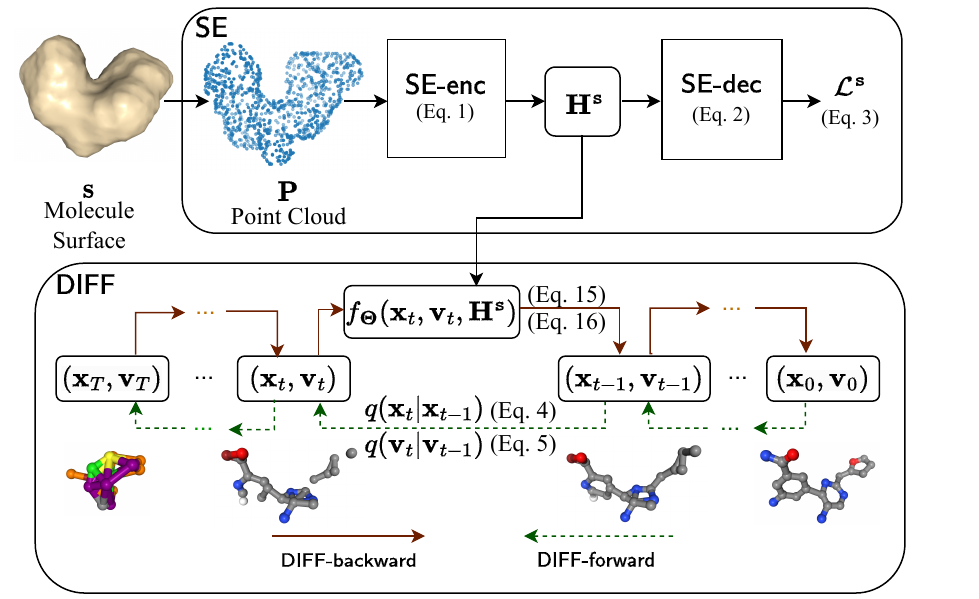}}
\vspace{-15pt}
\caption{Model Architecture of \method}
\label{fig:overall}
\vspace{-15pt}
\end{figure}

\method consists of an equivariant shape embedding module \SE that maps 3D molecular surface shapes to latent embeddings, and an equivariant diffusion model \methoddiff
that generates 3D molecules 
conditioned on these embeddings.
Figure~\ref{fig:overall} presents the overall architecture of \method. 
%

\subsection{Equivariant Shape Embedding (\SE)}
%
\method pre-trains a shape embedding module \SE to generate surface shape embeddings 
$\shapehiddenmat$. 
\SE uses an encoder \SEE to map $\pc$ to the equivariant latent embedding $\shapehiddenmat$.
\SE employs a decoder \SED to optimize $\shapehiddenmat$ by recovering the signed distances~\cite{park2019sdf} of sampled 
query points 
in 3D space to the molecule surface based on $\shapehiddenmat$. 
%
\method uses $\hiddenmat^{\mathtt{s}}$ to guide the diffusion process later. 
%


\subsubsection{Shape Encoder (\SEE)}

\SEE generates equivariant shape embeddings $\hiddenmat^{\mathtt{s}}$ from the 3D surface shape $\pc$ of molecules, 
such that $\hiddenmat^{\mathtt{s}}$ is equivariant to both translation and rotation of $\pc$.
That is, any translation and rotation applied to $\pc$ is reflected in $\hiddenmat^{\mathtt{s}}$ accordingly.
To ensure translation equivariance, 
\SEE shifts the center of each $\pc$ to zero to eliminate all translations.
To ensure rotation equivariance, 
\SEE leverages Vector Neurons (VNs)~\cite{deng2021vn} and Dynamic Graph Convolutional Neural Networks (DGCNNs)~\cite{wang2019dynamic} as follows:
\begin{equation*}
\{\hiddenmat^{\mathtt{p}}_1, \hiddenmat^{\mathtt{p}}_2, \cdots, \hiddenmat^{\mathtt{p}}_{|\scriptsize{\pc}|}\} = \text{VN-DGCNN}(\{\pointpos_1, \pointpos_2, \cdots, \pointpos_{|\scriptsize{\pc}|}\}), 
\vspace{-5pt}
\end{equation*}
\begin{equation*}
\hiddenmat^{\mathtt{s}} = \sum\nolimits_{j}\hiddenmat^{\mathtt{p}}_j / {|\pc|},
\label{eqn:shape_embed}
\end{equation*}
%
%
where $\text{VN-DGCNN}(\cdot)$ is a VN-based DGCNN network
to generate equivariant embedding $\hiddenmat^{\mathtt{p}}_j\in  \mathbb{R}^{d_p \times 3}$ for each point $z_j$ in $\pc$; 
and $\hiddenmat^{\mathtt{s}} \in \mathbb{R}^{d_p\times 3}$ is the embedding of $\pc$ generated via %
a mean-pooling over the embeddings of all the points.
Note that $\text{VN-DGCNN}(\cdot)$ generates a matrix as the embedding of each point (i.e., $\hiddenmat^{\mathtt{p}}_j$) to guarantee the equivariance.

\subsubsection{Shape Decoder (\SED)}

To optimize $\hiddenmat^{\mathtt{s}}$, \SE learns a decoder \SED to predict the signed distance of a query point $z_q$ sampled from  3D space using Multilayer Perceptrons (MLPs) as follows:
\begin{equation}
\tilde{o}_q = \text{MLP}(\text{concat}(\langle \mathbf{z}_q, \hiddenmat^{\mathtt{s}}\rangle, \|\mathbf{z}_q\|^2, \text{VN-In}(\hiddenmat^{\mathtt{s}}))),
\label{eqn:se:decoder}
\end{equation}
where $\tilde{o}_q$ is the predicted signed distance of $z_q$, with positive and negative values indicating $z_q$ is inside or outside the surface shape, respectively;
$\langle\cdot,\cdot\rangle$ is the dot-product operator;
$\|\mathbf{z}_q\|^2$ is the Euclidean norm of the coordinates of $z_q$;
$\text{VN-In}(\cdot)$ is an invariant 
VN network~\cite{deng2021vn} that converts the equivariant shape embedding $\shapehiddenmat\in \mathbb{R}^{d_p\times 3}$ into 
an invariant shape embedding. 
Thus, \SED predicts the signed distance between the query point and 3D surface by jointly considering the position of the query point ($\|\mathbf{z}_q\|^2$), the molecular surface shape ($\text{VN-In}(\cdot)$) and  the interaction between the point and surface $\langle\cdot,\cdot\rangle$.
The predicted signed distance $\tilde{o}_q$ is used to calculate the loss for the optimization of $\shapehiddenmat$ (discussed below).
As shown in the literature~\cite{deng2021vn}, $\tilde{o}_q$
remains invariant to the rotation of the 3D molecule surface shapes (i.e., $\pc$).
%
We present the sampling process of $z_q$ in the Supplementary Section~{\ref{supp:training:shapeemb}}.
%

\subsubsection{\SE Pre-training}

\method pre-trains \SE by minimizing the squared-errors loss between the predicted and the ground-truth signed distances of query points as follows:
\begin{equation}
\mathcal{L}^{\mathtt{s}} = \sum\nolimits_{z_q \in \mathcal{Z}} \|o_q-\tilde{o}_q\|^2,
\end{equation}
where $\mathcal{Z}$ is the set of sampled query points and $o_q$ is the ground-truth
signed distance of query point $z_q$.
By pretraining \SE, \method learns $\shapehiddenmat$ that will be used as the condition in the following 3D molecule generation.

%

\subsection{Shape-Conditioned Molecule Generation}
\label{section:diff}

In \method, a shape-conditioned molecule diffusion model, referred to as \methoddiff, 
is used to generate a 3D molecule structure (i.e., atom coordinates and features) 
conditioned on a given 3D surface shape 
that is represented by the shape latent embedding $\shapehiddenmat$ (Eq.~\ref{eqn:shape_embed}).
%
Following the denoising diffusion probabilistic models~\cite{ho2020ddpm}, \methoddiff includes a forward diffusion process based on a Markov chain,
denoted as \diffnoise, which gradually adds noises step by step to the atom positions and features $\{(\pos_i, \atomfeat_i)\}$ in the training molecules.
%
The noisy atom positions and features at step $t$ are represented as $\{(\pos_{i,t}, \atomfeat_{i,t})\}$ ($t=1, \cdots, T$),
and the molecules without any noise 
are represented as $\{(\pos_{i,0}, \atomfeat_{i,0})\}$.
At the final step $T$, $\{(\pos_{i,T}, \atomfeat_{i,T})\}$ are completely unstructured and resemble 
a simple distribution like a Normal distribution $\mathcal{N}(\mathbf{0}, \mathbf{I})$ or a uniform categorical distribution {$\mathcal{C}(\mathbf{1}/K)$},
in which $\mathbf{I}$ and $\mathbf{1}$ denotes the identity matrix and identity vector, respectively.

During training, \methoddiff is learned to reverse the forward diffusion process via another Markov chain, 
referred to as the backward generative process and denoted as \diffgenerative, to remove the noises in the noisy molecules.
During inference, 
\methoddiff first samples noisy atom positions and features at step $T$ 
from simple distributions and then generates a 3D molecule structure by removing the noises in the noisy molecules step by step until $t$ reaches 1.



\subsubsection{Forward Diffusion Process (\diffnoise)}
\label{section:diff:diff}

Following the previous work~\cite{guan2023targetdiff}, at step $t\in[1, T]$, a small Gaussian noise and a small categorical noise are added to the continuous atom positions and 
discrete atom features $\{(\pos_{i,t-1}, \atomfeat_{i,t-1})\}$, 
respectively.
When no ambiguity arises, we will eliminate subscript $i$ in the notations and use $(\pos_{t-1}, \atomfeat_{t-1})$ for brevity.
The noise levels of the Gaussian and categorical noises are determined by two predefined variance schedules $(\beta_t^{\mathtt{x}}, \beta_t^{\mathtt{v}})\in (0,1)$, where $\beta_t^{\mathtt{x}}$ and $\beta_t^{\mathtt{v}}$
are selected to be sufficiently small to ensure the smoothness of \diffnoise.
The details about variance schedules are available in Supplementary Section~\ref{supp:forward:variance}. 
%
%
Formally, for atom positions, the probability of $\pos_t$ sampled given $\pos_{t-1}$, denoted as $q(\pos_t|\pos_{t-1})$, is defined as follows,
%
\begin{equation}
q(\pos_t|\pos_{t-1}) = \mathcal{N}(\pos_t|\sqrt{1-\beta^{\mathtt{x}}_t}\pos_{t-1}, \beta^{\mathtt{x}}_t\mathbf{I}), 
\label{eqn:noiseposinter}
\end{equation}
%
where 
$\mathcal{N}(\cdot)$ is a Gaussian distribution of $\pos_t$ with mean $\sqrt{1-\beta_t^{\mathtt{x}}}\pos_{t-1}$ and covariance $\beta_t^{\mathtt{x}}\mathbf{I}$.
%
Following Hoogeboom \etal~\shortcite{hoogeboom2021catdiff}, 
%
for atom features, the probability of $\atomfeat_t$ across $K$ classes given $\atomfeat_{t-1}$ is defined as follows,
\begin{equation}
q(\atomfeat_t|\atomfeat_{t-1}) = \mathcal{C}(\atomfeat_t|(1-\beta^{\mathtt{v}}_t) \atomfeat_{t-1}+\beta^{\mathtt{v}}_t\mathbf{1}/K),
\label{eqn:noisetypeinter}
\end{equation}
where 
$\mathcal{C}$ is a categorical distribution of $\atomfeat_t$ derived by 
noising $\atomfeat_{t-1}$ with a uniform noise $\beta^{\mathtt{v}}_t\mathbf{1}/K$ across $K$ classes.

Since the above distributions form Markov chains, 
the probability of any $\pos_t$ or $\atomfeat_t$ can be derived from $\pos_0$ or $\atomfeat_0$:
%
\begin{eqnarray}
& q(\pos_t|\pos_{0}) & = \mathcal{N}(\pos_t|\sqrt{\cumalpha^{\mathtt{x}}_t}\pos_0, (1-\cumalpha^{\mathtt{x}}_t)\mathbf{I}), \label{eqn:noisepos}\\
& q(\atomfeat_t|\atomfeat_0)  & = \mathcal{C}(\atomfeat_t|\cumalpha^{\mathtt{v}}_t\atomfeat_0 + (1-\cumalpha^{\mathtt{v}}_t)\mathbf{1}/K), \label{eqn:noisetype}\\
& \text{where }\cumalpha^{\mathtt{u}}_t & = \prod\nolimits_{\tau=1}^{t}\alpha^{\mathtt{u}}_\tau, \ \alpha^{\mathtt{u}}_\tau=1 - \beta^{\mathtt{u}}_\tau, \ {\mathtt{u}}={\mathtt{x}} \text{ or } {\mathtt{v}}.\;\;\;\label{eqn:noiseschedule}
\label{eqn:pos_prior}
\end{eqnarray}
%
%
Note that $\bar{\alpha}^{\mathtt{u}}_t$ ($\mathtt{u}={\mathtt{x}}\text{ or }{\mathtt{v}}$)
is monotonically decreasing from 1 to 0 over $t=[1,T]$. 
As $t\rightarrow 1$, $\cumalpha^{\mathtt{x}}_t$ and $\cumalpha^{\mathtt{v}}_t$ are close to 1, leading to that $\pos_t$ or $\atomfeat_t$ approximates 
$\pos_0$ or $\atomfeat_0$.
Conversely, as  $t\rightarrow T$, $\cumalpha^{\mathtt{x}}_t$ and $\cumalpha^{\mathtt{v}}_t$ are close to 0,
leading to that $q(\pos_T|\pos_{0})$ 
resembles  {$\mathcal{N}(\mathbf{0}, \mathbf{I})$} 
and $q(\atomfeat_T|\atomfeat_0)$ 
resembles {$\mathcal{C}(\mathbf{1}/K)$}.

Using Bayes theorem, the ground-truth Normal posterior of atom positions $p(\pos_{t-1}|\pos_t, \pos_0)$ can be calculated in a
closed-form~\cite{ho2020ddpm} as below,
\begin{eqnarray}
& p(\pos_{t-1}|\pos_t, \pos_0) = \mathcal{N}(\pos_{t-1}|\mu(\pos_t, \pos_0), \tilde{\beta}^\mathtt{x}_t\mathbf{I}), \label{eqn:gt_pos_posterior_1}\\
&\!\!\!\!\!\!\!\!\!\!\!\mu(\pos_t, \pos_0)\!=\!\frac{\sqrt{\bar{\alpha}^{\mathtt{x}}_{t-1}}\beta^{\mathtt{x}}_t}{1-\bar{\alpha}^{\mathtt{x}}_t}\pos_0\!+\!\frac{\sqrt{\alpha^{\mathtt{x}}_t}(1-\bar{\alpha}^{\mathtt{x}}_{t-1})}{1-\bar{\alpha}^{\mathtt{x}}_t}\pos_t, 
 \tilde{\beta}^\mathtt{x}_t\!=\!\frac{1-\bar{\alpha}^{\mathtt{x}}_{t-1}}{1-\bar{\alpha}^{\mathtt{x}}_{t}}\beta^{\mathtt{x}}_t.\;\;\;
\end{eqnarray}
%
Similarly, the ground-truth categorical posterior of atom features $p(\atomfeat_{t-1}|\atomfeat_{t}, \atomfeat_0)$ can be calculated~\cite{hoogeboom2021catdiff} as below,
\begin{eqnarray}
& p(\atomfeat_{t-1}|\atomfeat_{t}, \atomfeat_0) = \mathcal{C}(\atomfeat_{t-1}|\mathbf{c}(\atomfeat_t, \atomfeat_0)), \label{eqn:gt_atomfeat_posterior_1}\\
& \mathbf{c}(\atomfeat_t, \atomfeat_0) = \tilde{\mathbf{c}}/{\sum_{k=1}^K \tilde{c}_k}, \label{eqn:gt_atomfeat_posterior_2} \\
& \tilde{\mathbf{c}} = [\alpha^{\mathtt{v}}_t\atomfeat_t + \frac{1 - \alpha^{\mathtt{v}}_t}{K}]\odot[\bar{\alpha}^{\mathtt{v}}_{t-1}\atomfeat_{0}+\frac{1-\bar{\alpha}^{\mathtt{v}}_{t-1}}{K}], 
\label{eqn:gt_atomfeat_posterior_3}
\end{eqnarray}
%
%
where $\tilde{c}_k$ denotes the likelihood of $k$-th class across $K$ classes in $\tilde{\mathbf{c}}$; 
$\odot$ denotes the element-wise product operation;
$\tilde{\mathbf{c}}$ is calculated using $\atomfeat_t$ and $\atomfeat_{0}$ and normalized so as to represent
probabilities. 
%
The proof of the above equations is available in Supplementary Section~\ref{supp:forward:proof}.

\subsubsection{{Backward Generative Process (\diffgenerative)}}
\label{section:diff:backward}

\methoddiff learns to reverse {\diffnoise} by denoising from {$(\pos_{t}, \atomfeat_{t})$} to {$(\pos_{t-1}, \atomfeat_{t-1})$} 
at $t\in[1,T]$, conditioned on the shape latent embedding $\shapehiddenmat$.
Specifically, the probabilities of $(\pos_{t-1}, \atomfeat_{t-1})$ denoised from $(\pos_{t}, \atomfeat_{t})$ are estimated by 
the approximates of the ground-truth posteriors $p(\pos_{t-1}|\pos_t, \pos_0)$ (Eq.~\ref{eqn:gt_pos_posterior_1}) and 
$p(\atomfeat_{t-1}|\atomfeat_{t}, \atomfeat_0)$ (Eq.~\ref{eqn:gt_atomfeat_posterior_1}).
%
Given that $(\pos_0, \atomfeat_0)$ is unknown in the generative process,
a predictor $f_{\boldsymbol{\Theta}}(\pos_t, \atomfeat_t, \shapehiddenmat)$ is employed to predict at  $t$
the atom position and feature $(\pos_0, \atomfeat_0)$
as below,
\begin{equation}
(\tilde{\pos}_{0,t}, \tilde{\atomfeat}_{0,t}) = f_{\boldsymbol{\Theta}}(\pos_t, \atomfeat_t, \hiddenmat^{\mathtt{s}}),
\label{eqn:predictor}
\end{equation}
where $\tilde{\pos}_{0,t}$ and $\tilde{\atomfeat}_{0,t}$ are the predictions of $\pos_0$ and $\atomfeat_0$ at $t$;
${\boldsymbol{\Theta}}$
is the learnable parameter.
Following Ho \etal~\shortcite{ho2020ddpm}, with $\tilde{\pos}_{0,t}$, the probability of $\pos_{t-1}$ denoised from $\pos_t$, denoted as $p(\pos_{t-1}|\pos_t)$,
can be estimated 
by the approximated posterior $p_{\boldsymbol{\Theta}}((\pos_{t-1}|\pos_t, \tilde{\pos}_{0,t})$ as below,
\begin{equation}
\begin{aligned}
p(\pos_{t-1}|\pos_t) & \approx p_{\boldsymbol{\Theta}}(\pos_{t-1}|\pos_t, \tilde{\pos}_{0,t}) \\
& = \mathcal{N}(\pos_{t-1}|\mu_{\boldsymbol{\Theta}}(\pos_t, \tilde{\pos}_{0,t}), \tilde{\beta}^\mathtt{x}_t\mathbf{I}),
\end{aligned}
\label{eqn:aprox_pos_posterior}
\end{equation}
where $\mu_{\boldsymbol{\Theta}}(\pos_t, \tilde{\pos}_{0,t})$ is an estimate 
of $\mu(\pos_t, \pos_{0})$ by replacing $\pos_0$ with its estimate $\tilde{\pos}_{0,t}$ 
in Eq.~{\ref{eqn:gt_pos_posterior_1}}.
Similarly, with $\tilde{\atomfeat}_{0,t}$, the probability of $\atomfeat_{t-1}$ denoised from $\atomfeat_t$, denoted as $p(\atomfeat_{t-1}|\atomfeat_t)$, 
can be estimated 
by the approximated posterior $p_{\boldsymbol{\Theta}}(\atomfeat_{t-1}|\atomfeat_t, \tilde{\atomfeat}_{0,t})$ as below,
\begin{equation}
\begin{aligned}
\!\!\!\!\!\!\!\!p(\atomfeat_{t-1}|\atomfeat_t)\!\!\approx\!\! p_{\boldsymbol{\Theta}}(\atomfeat_{t-1}|\atomfeat_{t}, \tilde{\atomfeat}_{0,t}) 
\!\!=\!\!\mathcal{C}(\atomfeat_{t-1}|\mathbf{c}_{\boldsymbol{\Theta}}(\atomfeat_t, \tilde{\atomfeat}_{0,t})),\!\!\!\!
\end{aligned}
\label{eqn:aprox_atomfeat_posterior}
\end{equation}
where $\mathbf{c}_{\boldsymbol{\Theta}}(\atomfeat_t, \tilde{\atomfeat}_{0,t})$ is an estimate of $\mathbf{c}(\atomfeat_t, \atomfeat_0)$
by replacing $\atomfeat_0$  
with its estimate $\tilde{\atomfeat}_{0,t}$ in Eq.~\ref{eqn:gt_atomfeat_posterior_1}.

\subsubsection{{Equivariant Shape-Conditioned Molecule Predictor}}
\label{section:diff:nn}

In \diffgenerative, the predictor $f_{\boldsymbol{\Theta}}(\pos_t, \atomfeat_t, \shapehiddenmat)$ (Eq.~\ref{eqn:predictor}) 
predicts the atom positions and features $(\tilde{\pos}_{0,t}, \tilde{\atomfeat}_{0,t})$ 
given the noisy data $(\pos_t, \atomfeat_t)$ conditioned on 
$\shapehiddenmat$.
For brevity, in this subsection, we eliminate the subscript $t$ in the notations when no ambiguity arises.
$f_{\boldsymbol{\Theta}}(\cdot)$ leverages two multi-layer graph neural networks: 
(1) an equivariant graph neural network, denoted as \eqgnn, that equivariantly predicts atom positions that change under transformations, and 
(2) an invariant graph neural network, denoted as \invgnn, that invariantly predicts atom features that remain unchanged under transformations.
%
%
%
Following the previous work~\cite{guan2023targetdiff,hoogeboom22diff}, the translation equivariance of atom position prediction is achieved by 
shifting a fixed point (e.g., the center of point clouds \pc) to zero, and therefore only rotation equivariance needs to be considered.
%

\paragraph{Atom Coordinate Prediction}

In \eqgnn, 
the atom position $\pos_{i}^{l+1}\in \mathbb{R}^{3}$ of $a_i$ 
 at the ($l$+1)-th layer is calculated in an equivariant way as below,
\begin{equation*}
 \Delta\pos_i^{l+1} \! = \sum_{\mathclap{j\in N(a_i), i\neq j}}(\pos_i^l - \pos_j^l)\text{MHA}^{\mathtt{x}}(d_{ij}^{l}, \hiddenvec_i^{l+1}, \hiddenvec_j^{l+1}, \text{VN-In}(\shapehiddenmat)),
\end{equation*}
\vspace{-5pt}
\begin{equation} 
\label{eqn:diff:graph:pos}
 \pos_i^{l+1} \! = \pos_i^{l}\! + \!\text{Mean}(\Delta{\pos_{i}^{l+1}})\! + \!\text{VN-Lin}(\pos_i^l, \Delta\pos_i^{l+1}, \shapehiddenmat),
\end{equation}
%
where $N(a_i)$ is the set of $N$-nearest neighbors of $a_i$ 
based on atomic distances; 
$\Delta\pos_i^{l+1}\in \mathbb{R}^{n_h\times 3}$  
aggregates the neighborhood information of $a_i$; 
$\text{MHA}^{\mathtt{x}}(\cdot)$ denotes the multi-head attention layer in \eqgnn 
with $n_h$ heads;
$d_{ij}^l$ is the distance between $i$-th and $j$-th atom positions $\pos_i^{l}$ and $\pos_j^{l}$ at the $l$-th layer;
$\text{Mean}(\Delta{\pos_{i}^{l+1}})$ converts $\Delta\pos_i^{l+1}$ into a 3D vector via meaning pooling
to adjust the atom position; 
$\text{VN-Lin}(\cdot)\in \mathbb{R}^{3}$ 
denotes the equivariant VN-based linear layer~\cite{deng2021vn}. 
%
%
%
$\text{VN-Lin}(\cdot)$ adjusts the atom positions to fit the shape condition represented by $\shapehiddenmat$, by considering the 
current atom positions $\pos_i^l$ and the neighborhood information $\Delta\pos_i^{l+1}$.
The learned atom position $\pos_i^{L}$ at the last layer $L$ of \eqgnn is used as the prediction of 
$\tilde{\pos}_{i, 0}$, that is,  
\begin{equation}
\tilde{\pos}_{i, 0} = \pos_i^{L}. 
\end{equation}

\paragraph{Atom Feature Prediction}

In \invgnn, inspired by the previous work~\cite{guan2023targetdiff} and VN-Layer~\cite{deng2021vn},
the atom feature embedding $\hiddenvec_{i}^{l+1}\in \mathbb{R}^{d_h}$ of the $i$-th atom $a_i$ at the ($l$+1)-th layer of \invgnn
is updated in an invariant way as follows,
\begin{equation}
\label{eqn:diff:graph:atomfeat}
\!\!\!\hiddenvec_{i}^{l+1}\!\!=\!\!\hiddenvec_{i}^l\!+\!\sum_{\mathclap{j\in N(a_i), i\neq j}} \text{MHA}^{\mathtt{h}}(d_{ij}^l, \hiddenvec_i^l, \hiddenvec_j^l, \text{VN-In}(\shapehiddenmat)),\\
\hiddenvec_{i}^0\!=\!\atomfeat_i, \!\!\!
\end{equation}
where $\text{MHA}^{\mathtt{h}}(\cdot)\in\mathbb{R}^{d_h}$ denotes the multi-head attention layer in \invgnn. 
%
The learned atom feature embedding $\hiddenvec_{i}^{L}$ at the last layer $L$ encodes the neighborhood information of $a_i$ and the conditioned molecular shape, and is used to predict the atom features as follows: 
%
\begin{equation}
\tilde{\atomfeat}_{i, 0} = \text{softmax}(\text{MLP}(\hiddenvec_{i}^{L})).
\label{eqn:diff:graph:atompred}
\end{equation}
The proof of equivariance in Eq.~\ref{eqn:diff:graph:pos}
and invariance in Eq.~\ref{eqn:diff:graph:atomfeat} is available in Supplementary Section~\ref{supp:backward:equivariance} and ~\ref{supp:backward:invariance}. 

\subsubsection{Model Training}
\label{section:diff:opt}

\method optimizes \methoddiff by minimizing the squared errors between the predicted positions ($\tilde{\pos}_{0,t}$) and the ground-truth positions ($\pos_0$) of atoms in molecules.
Given a particular step $t$, the error is calculated as follows:
\begin{equation}
\label{eqn:diff:obj:pos}
\begin{aligned}
& \mathcal{L}^\mathtt{x}_t({\mol}) = w_t^\mathtt{x}\sum\nolimits_{\forall a \in {\scriptsize{\mol}}}\|\tilde{\pos}_{0,t} - \pos_{0}\|^2, \\
& \text{where} \ w_t^\mathtt{x} = \min(\lambda_t, \delta),\ \lambda_t={\bar{\alpha}^{\mathtt{x}}_t}/({1-\bar{\alpha}^{\mathtt{x}}_t}),
\end{aligned}
\end{equation}
%
where $w_t^\mathtt{x}$ is a weight at step $t$, and is calculated by clipping the signal-to-noise ratio 
$\lambda_t>0$ 
with a threshold $\delta > 0$. 
Note that because $\bar{\alpha}_t^{\mathtt{x}}$ decreases monotonically as $t$ increases from 1 to $T$ (Eq.~\ref{eqn:noiseschedule}), $w_t^\mathtt{x}$ decreases monotonically over $t$ as well until it is clipped. 
Thus, $w_t^\mathtt{x}$ imposes lower weights on the loss when the noise level in $\mathtt{x}_t$ is higher (i.e., at later/larger step $t$). 
This encourages the model training to focus more on accurately recovering molecule structures when there are 
sufficient signals in the data, rather than being potentially confused by major noises in the data. 


\method also minimizes the KL divergence~\cite{kullback1951information} between the 
ground-truth posterior $p(\atomfeat_{t-1}|\atomfeat_t, \atomfeat_0)$ (Eq.~\ref{eqn:gt_atomfeat_posterior_1}) 
and its approximate
$p_\theta(\atomfeat_{t-1}|\atomfeat_{t}, \tilde{\atomfeat}_{0,t})$ (Eq.~\ref{eqn:aprox_atomfeat_posterior}) 
for discrete atom features
to optimize \methoddiff, following the literature~\cite{hoogeboom2021catdiff}. 
%
Particularly, the KL divergence at $t$ for a given molecule is calculated as follows:
%
\begin{equation*}
\mathcal{L}^{\mathtt{v}}_t({\mol})  = \sum\nolimits_{\forall a \in \scriptsize{\mol}}\text{KL}(p(\atomfeat_{t-1}|\atomfeat_{t}, \atomfeat_{0}) | p_{\boldsymbol{\Theta}}(\atomfeat_{t-1}|\atomfeat_{t}, \tilde{\atomfeat}_{0,t}) ),
\end{equation*}
\vspace{-10pt}
\begin{equation}
\label{eqn:lossv}
\! \! \! \! \! \!\! \! \! \! \! \!= \sum\nolimits_{\forall a\in \scriptsize{\mol}}\text{KL}(\mathbf{c}(\atomfeat_{t}, \atomfeat_{0})|\mathbf{c}_{\boldsymbol{\Theta}}(\atomfeat_{t}, \tilde{\atomfeat}_{0,t})),
\end{equation}
where $\mathbf{c}(\atomfeat_{t}, \atomfeat_{0})$ is a categorical distribution of $\atomfeat_{t-1}$  (Eq.~\ref{eqn:gt_atomfeat_posterior_2});
$\mathbf{c}_{\boldsymbol{\Theta}}(\atomfeat_{t}, \tilde{\atomfeat}_{0,t})$ is an estimate of $\mathbf{c}(\atomfeat_{t}, \atomfeat_{0})$ (Eq.~\ref{eqn:aprox_atomfeat_posterior}).
%
%
The overall \method loss function  is defined as follows:
\begin{equation}
\label{eqn:loss}
\mathcal{L} =\sum\nolimits_{\forall \scriptsize{\mol}\in\mathcal{M}}\sum\nolimits_{\forall t \in \mathcal{T}} (\mathcal{L}^{\mathtt{x}}_t(\mol) + \xi \mathcal{L}^{\mathtt{v}}_t(\mol)),
\end{equation}
where $\mathcal{M}$ is the set of all the molecules in training; 
$\mathcal{T}$ is the set of sampled timesteps; 
$\xi > 0$ is a hyper-parameter to balance $\mathcal{L}^{\mathtt{x}}_t$(\mol) and $\mathcal{L}^{\mathtt{v}}_t$(\mol).
During training, step $t$ is uniformly sampled from $\{1, 2, \cdots, 1000\}$.
The derivation of the loss functions
is available in Supplementary Section~\ref{supp:training:loss}. 

\subsubsection{Molecule Generation}
\label{section:diff:sample}

During inference, \method generates novel molecules by gradually denoising $(\pos_T, \atomfeat_T)$ to $(\pos_0, \atomfeat_0)$ using the equivariant shape-conditioned molecule predictor.
Specifically, \method samples $\pos_T$ and $\atomfeat_T$ from $\mathcal{N}(\mathbf{0}, \mathbf{I})$ and $\mathcal{C}(\mathbf{1}/K)$, respectively.
After that, \method samples $\pos_{t-1}$ from $\pos_t$ using $p_{\boldsymbol{\Theta}}(\pos_{t-1}|\pos_t, \tilde{\pos}_{0,t})$ (Eq.~\ref{eqn:aprox_pos_posterior}).
Similarly, \method samples $\atomfeat_{t-1}$ from $\atomfeat_{t}$ using $p_{\boldsymbol{\Theta}}(\atomfeat_{t-1}|\atomfeat_{t}, \tilde{\atomfeat}_{0,t})$ (Eq.~\ref{eqn:aprox_atomfeat_posterior}) until $t$ reaches 1.

\subsubsection{{\method with Shape Guidance}}
\label{section:diff:shape-guide}

During molecule generation, \method can also utilize additional shape guidance
by pushing the predicted atoms to the shape 
of the given molecule \molx.
Particularly, following Adams and Coley \etal~\shortcite{adams2023equivariant}, 
the shape used for guidance is defined as a set of points $\mathcal{Q}$ sampled according to atom positions in \molx.
Particularly, for each atom $a_i$ in \molx, 20 points are randomly sampled into $\mathcal{Q}$ from a 
Gaussian distribution centered at $\pos_i$ with variance $\phi$. 
%
Given the predicted atom position $\tilde{\pos}_{0,t}$ at step $t$, \method applies the shape guidance 
by adjusting the predicted positions to \molx as follows:
\begin{equation}
\label{eqn:guidance}
\begin{aligned}
\!\!\!\!\!\!\!\!\!\!\!\!\!\!\!\pos_{0,t}^*\!\!=\!\!(1\!-\!\sigma) \tilde{\pos}_{0,t}\!\!+\!\sigma \!{\sum_{\mathclap{\mathbf{z}\in n(\tilde{\scriptsize{\pos}}_{0,t}; \mathcal{Q})}} \mathbf{z}}/{n},
 \text{when }\!\!\sum_{\mathclap{\mathbf{z}\in n(\tilde{\scriptsize{\pos}}_{0,t}; \mathcal{Q})}} d(\tilde{\pos}_{0,t}, \mathbf{z}) / n\!>\!\gamma,\!\!\!\!\!\!\!\!\!\!\!\!\!
\end{aligned}
\end{equation}
where $\sigma>0$ is the weight used to balance the prediction $\tilde{\pos}_{0,t}$ and the adjustment;
$d(\tilde{\pos}_{0, t}, \mathbf{z})$ is the Euclidean distance between $\tilde{\pos}_{0, t}$ and $\mathbf{z}$;
$n(\tilde{\pos}_{0,t};\mathcal{Q})$ is the set of $n$-nearest neighbors of $\tilde{\pos}_{0,t}$ in $\mathcal{Q}$ 
based on $d(\cdot)$;
$\gamma>0$ is a distance threshold. 
By doing the above adjustment, the predicted atom positions will be pushed to those of {\molx} if they
are sufficiently far away. 
Note that the shape guidance is applied exclusively for steps 
\begin{equation}
\label{eqn:steps}
t=T, T-1, \cdots, S\text{, where } S>1, 
\end{equation}
not for all the steps, 
and thus it only adjusts predicted atom positions when there are a lot of noises and the prediction needs 
more guidance. 
%
%
\method with the shape guidance is referred to as \methodwithguide.
\section{Experiments}

\vspace{-3pt}

\subsection{Data}

Following \squid~\cite{adams2023equivariant}, we used molecules in the MOSES dataset~\cite{mose2020}, with their 3D conformers calculated by RDKit~\cite{rdkit}.
We used the same training and test split as in \squid.
%
%
Please note that \squid further modifies the generated conformers into artificial ones, by adjusting acyclic bond distances to their empirical means 
and fixing acyclic bond angles using heuristic rules.
Unlike \squid, we did not make any additional adjustments to the calculated 3D conformers, as \method is designed with sufficient flexibility to accept any 3D conformers as input
and generate 3D molecules without restrictions on fixed bond lengths or angles. 
%
Limited by the predefined fragment library, \squid also removes molecules with fragments not present in its fragment library. 
In contrast, we kept all the molecules, as {\method} is not based on fragments. 
Our final training dataset contains 1,593,653 molecules, out of which a random set of 1,000 molecules was selected for validation.
Both the {\methodenc} and {\methoddiff} models are trained using this training set.
1,000 test molecules (i.e., conditions) as used in \squid are used to test \method. 

\subsection{Baselines}

We compared \method and \methodwithguide with the state-of-the-art baseline \squid and a virtual screening method over the training dataset, 
denoted as \dataset.
%
As far as we know, \squid is the only generative baseline that 
generates 3D molecules conditioned on molecule shapes.
%
%
%
\squid consists of a fragment-based generative model based on variational autoencoder that sequentially decodes fragments from molecule latent embeddings and shape embeddings,
and a rotatable bond scoring framework that adjusts the angles of rotatable bonds between fragments to maximize the 3D shape similarity with the condition molecule.
%
%
\dataset aims to sift through the training set to identify molecules with high shape similarities with the condition
molecule.
For \squid, we assessed two interpolation levels, $\lambda=0.3$ and $1.0$ (prior), 
following the original \squid paper~\cite{adams2023equivariant}. 
For {\squid}, {\method} and {\methodwithguide}, we generated 50 molecules for each testing molecule (i.e., condition) as the candidates for evaluation.
For {\dataset}, we randomly sampled 500 training molecules for each testing molecule, and considered the top-50 molecules with the highest shape similarities as candidates for evaluation.

%

\subsubsection{Evaluation Metrics}

We use shape similarity $\shapesim(\shape_x, \shape_y)$ and molecular graph similarity 
$\graphsim(\molx, \moly)$ to measure the generated new molecules \moly with respective to 
the condition \molx. Higher $\shapesim$ and meanwhile lower \graphsim indicate better 
model performance. 
We also measure the diversity (\diversity) of the generated molecules, calculated as 1 minus
average pairwise \graphsim among all generated molecules. 
Higher \diversity \ indicates better performance.  
Details about the evaluation metrics are available in Supplementary Section~\ref{supp:experiments:metrics}. 

\vspace{-5pt}
\subsection{Performance Comparison}

\begin{table*}
	\centering
		\caption{{Overall Comparison on Shape-Conditioned Molecule Generation}}
	\label{tbl:overall}
	\begin{small}
\begin{threeparttable}
	\begin{tabular}{
		@{\hspace{0pt}}l@{\hspace{2pt}}
		@{\hspace{5pt}}r@{\hspace{5pt}}
		@{\hspace{5pt}}r@{\hspace{5pt}}
		@{\hspace{5pt}}r@{\hspace{5pt}}
		@{\hspace{5pt}}r@{\hspace{5pt}}
		@{\hspace{5pt}}r@{\hspace{5pt}}
		@{\hspace{0pt}}c@{\hspace{0pt}}
	    	@{\hspace{5pt}}c@{\hspace{5pt}}
		@{\hspace{5pt}}c@{\hspace{5pt}}
		@{\hspace{5pt}}c@{\hspace{3pt}}
		}
		\toprule
		method & \#c\% & \#u\%  & \qed  & \avgshapesim(std) & \avggraphsim(std) & & \maxshapesim(std) & \maxgraphsim(std) &  \diversity(std) \\
		\midrule
		\dataset                                & 100.0 & 100.0 & 0.795 & 0.729 (0.039) & 0.226 (0.038) & & 0.807 (0.042) & 0.241 (0.087) & 0.759 (0.015)\\
		\squid ($\lambda$=0.3)        & 100.0 &   94.2 & 0.766 & 0.717 (0.083) & 0.349 (0.088) & & 0.904 (0.070) & 0.549 (0.243) & 0.677 (0.065) \\
		\squid ($\lambda$=1.0)        & 100.0 &   95.0 & 0.760 & 0.670 (0.069) & 0.235 (0.045) & & 0.842 (0.061) & 0.271 (0.096) & 0.744 (0.046) \\
		\method                                & 98.8 & 100.0 & 0.748 & 0.689 (0.044) & 0.239 (0.049) & & 0.803 (0.042) & 0.243 (0.068) & 0.712 (0.055) \\
		\methodwithguide                 & 98.7 & 100.0 & 0.749 & 0.746 (0.036) & 0.241 (0.050) & & 0.852 (0.034) & 0.247 (0.068) & 0.703 (0.053) \\
		%
		\bottomrule
	\end{tabular}%
	\begin{tablenotes}[normal,flushleft]
		\begin{footnotesize}
	\item 
\!\!Columns represent: ``\#c\%": the percentage of connected molecules; ``\#u\%'': the percentage of unique molecules;
``\qed'': the average drug-likeness of generated molecules;
``\avgshapesim/\avggraphsim'': the average of shape or graph similarities between the condition molecules and generated molecules; ``std": the standard deviation;
``\maxshapesim'': the maximum of shape similarities between the condition molecules and generated molecules;
``\maxgraphsim'': the graph similarities between the condition molecules and the molecules with the maximum shape similarities;
``\diversity'': the diversity among the generated molecules. \par
		\par
		\end{footnotesize}
	\end{tablenotes}
\end{threeparttable}
\end{small}
  \vspace{-10pt}    
\end{table*}


\subsubsection{Overall Comparison}

Table~\ref{tbl:overall} presents the overall comparison of shape-conditioned molecule generation among 
\dataset, \squid, \method and \methodwithguide. 
%
%
%
As shown in Table~\ref{tbl:overall}, \methodwithguide achieves the highest average shape similarity 0.746$\pm$0.036,
with 2.3\% improvement from the best baseline \dataset (0.729$\pm$0.039), although at the cost of a slightly higher graph similarity (0.241$\pm$0.050 in \methodwithguide vs 0.226$\pm$0.038 in \dataset ).
%
%
This indicates that \methodwithguide could generate molecules that align more closely with the shape conditions than those in the dataset.
Furthermore, \methodwithguide achieves the second-best performance in maximum shape similarity
\maxshapesim \ at 0.852$\pm$0.034 among all the methods.
While it underperforms the best baseline (0.904$\pm$0.070 for \squid with $\lambda$=0.3) on this metric, \methodwithguide achieves substantially 
lower maximum graph similarity \maxgraphsim \ of 0.247$\pm$0.068 compared with the best baseline (0.549$\pm$0.243).
This highlights the ability of \methodwithguide in generating novel molecules that resemble the shape conditions.
\methodwithguide also achieves the lowest standard deviation values on both the average and maximum shape similarities (0.036 and 0.034, respectively) among all the methods,
further demonstrating its ability to consistently generate molecules with high shape similarities.


\methodwithguide performs substantially better than \method on 3D shape similarity metrics (e.g., 0.746$\pm$0.036 vs 0.689$\pm$0.044 on \avgshapesim).
%
The superior performance of \methodwithguide highlights the importance of shape guidance in the generative process.
Although \method underperforms \methodwithguide, it still outperforms \squid with $\lambda$=1.0 in terms of the \avgshapesim \ (i.e., 0.689$\pm$0.044 vs 0.670$\pm$0.069).

In terms of the quality of generated molecules, 98.7\% of molecules from \methodwithguide and 98.8\% from \method are connected, and every connected molecule is unique.
\squid with $\lambda$ values of 0.3 or 1.0 ensures the 100\% connectivity among generated molecules by sequentially attaching fragments.
However, out of these connected molecules, 94.2\% and 95.0\% are unique for \squid with $\lambda$ value of 0.3 or 1.0, respectively.
In terms of the drug-likeness (QED), both \methodwithguide and \method achieve QED values (e.g., 0.749 for \methodwithguide) close to those of \squid with $\lambda$ as 0.3 and 1.0 (e.g., 
0.760 for \squid with $\lambda$=0.3).
All the generative methods produce slightly inferior QED values to real molecules (0.795 for \dataset).
In terms of diversity, \methodwithguide and \method achieve higher diversity values (e.g., 0.703$\pm$0.053 for \methodwithguide) than \squid with $\lambda$=1.0 (0.677$\pm$0.065), though 
slightly lower than \squid with $\lambda$=0.3 and \dataset.
Overall, \method and \methodwithguide are able to generate connected, unique and diverse molecules with good drug-likeness scores.

Please note that unlike \squid, which neglects distorted bonding geometries in real molecules and limits itself to generating molecules with fixed bond lengths and angles, both \method and \methodwithguide are able to generate molecules without such limitations. 
Given the superior performance of \methodwithguide in shape-conditioned molecule generation, it could serve as a promising tool for ligand-based drug design.
%

\subsubsection{Comparison of Diffusion Weighting Schemes}

%

\begin{table}
	\centering
		\caption{Comparison of Diffusion Weighting Schemes}
	\label{tbl:ablation_study:timestepweight}
	\begin{small}
\begin{threeparttable}
	\begin{tabular}{
		@{\hspace{3pt}}l@{\hspace{0pt}}
		@{\hspace{3pt}}c@{\hspace{3pt}}
		@{\hspace{3pt}}r@{\hspace{3pt}}
		@{\hspace{3pt}}r@{\hspace{3pt}}
		@{\hspace{3pt}}r@{\hspace{3pt}}
		@{\hspace{6pt}}r@{\hspace{3pt}}
		%
		%
		%
		@{\hspace{3pt}}c@{\hspace{3pt}}
		}
		\toprule
		\multirow{2}{*}{method} & \multirow{2}{*}{weights} & \multirow{2}{*}{\#c\%} & \multirow{2}{*}{\#u\%} &\multirow{2}{*}{\qed} & \multicolumn{2}{c}{JS divergence}\\
		\cmidrule(r){6-7}
		& & & &  & bond & C-C\\
		\midrule
		\multirow{2}{*}{\method}   & $w^{\mathtt{x}}_t$ & 98.8 & 100.0 & 0.748 & 
		 0.095 &  0.321 \\
		&   uniform   & 89.4 & 100.0 & 0.660 & 
		0.115 & 0.393 \\
		\midrule
		\multirow{2}{*}{\methodwithguide} & $w^{\mathtt{x}}_t$  & 98.7 & 100.0 & 0.749 & 
		0.093 & 0.317 \\
		&   uniform   & 90.1 & 100.0 & 0.671 & 
		0.112 & 0.384 \\
		\bottomrule
	\end{tabular}%
	\begin{tablenotes}[normal,flushleft]
		\begin{footnotesize}
	\item 
\!\!Columns represent: 
``weights": different weighting schemes;
``JS distance of bond/C-C'': the Jensen-Shannon (JS) divergence of bond length among all the bond types (``bond")/carbon-carbon single bonds (``C-C") between real molecules and generated molecules; 
All the others are identical to those in Table~\ref{tbl:overall}. 
		\par
		\end{footnotesize}
	\end{tablenotes}
\end{threeparttable}
\end{small}
 \vspace{-10pt}    
\end{table}


%
While previous work~\cite{peng2023moldiff, guan2023targetdiff} applied uniform weights on different  
diffusion steps, 
\method uses different weights (i.e., $w^{\mathtt{x}}_t$ in Eq.~\ref{eqn:diff:obj:pos}). 
%
We conducted an ablation study to demonstrate the effectiveness of this new weighting scheme. 
Particularly, we trained two \methoddiff modules
with the varying step weights $w^{\mathtt{x}}_t$ (with $\delta=10$ in Eq.~\ref{eqn:diff:obj:pos})
and uniform weights, respectively, while fixing all the other hyper-parameters in \method and \methodwithguide.
Table~\ref{tbl:ablation_study:timestepweight} presents their performance comparison. 
%

The results in Table~\ref{tbl:ablation_study:timestepweight} show that the different weights on different steps substantially improve the quality of the generated molecules.
Specifically, \method with different weights ensures higher molecular connectivity and drug-likeness than that 
with uniform weights ($98.8\%$ vs $89.4\%$ for connectivity; $0.748$ vs $0.660$ for QED).
%
%
\method with different weights also produces molecules with bond length distributions closer to those of real molecules (i.e., lower Jensen-Shannon divergence),
for example, the Jensen-Shannon (JS) divergence of bond lengths between real and generated molecules
decreases from 0.115 to 0.095 when different weights are applied.
The same trend can be observed for \methodwithguide, for which the different weights also improve the 
generated molecule qualities. 
Since $w^{\mathtt{x}}_t$ increases as the noise level in the data decreases (See discussions earlier in 
``Model Training"), the results in Table~{\ref{tbl:ablation_study:timestepweight}} demonstrate the effectiveness
of the new weighting scheme in promoting new molecules generated more similarly to real ones
when the noise level in data is small. 


%
%

\subsubsection{Parameter Study}

\begin{table}[!t]
	\centering
		\caption{{Parameter Study in Shape Guidance}}
	\label{tbl:ablation_study:shape_guidance}
	\begin{small}
\begin{threeparttable}
	\begin{tabular}{
		@{\hspace{2pt}}r@{\hspace{3pt}}
		@{\hspace{3pt}}r@{\hspace{3pt}}
		@{\hspace{2pt}}r@{\hspace{2pt}}
		@{\hspace{2pt}}r@{\hspace{2pt}}
		@{\hspace{2pt}}r@{\hspace{2pt}}
		@{\hspace{0pt}}r@{\hspace{0pt}}
		@{\hspace{2pt}}r@{\hspace{0pt}}
		@{\hspace{0pt}}r@{\hspace{2pt}}
		@{\hspace{0pt}}c@{\hspace{0pt}}
	    	@{\hspace{2pt}}c@{\hspace{2pt}}
		@{\hspace{2pt}}c@{\hspace{2pt}}
		}
		\toprule
		$\gamma$ & $S$ & \qed & JS. bond & & \avgshapesim & \avggraphsim  & & \maxshapesim & \maxgraphsim  \\
		\midrule
		- & - &  0.748 & 
		0.094  & & 0.689  & 0.239  & & 0.803 & 0.243 \\
		\midrule
		0.2 & 50 &  0.630 & 
		0.110 & & 0.794 & 0.236 & & 0.890 & 0.244 \\
		0.2 & 100 &  0.666 & 
		0.105 & & 0.786  & 0.238  & & 0.883  & 0.245  \\
		0.2 & 300 &  0.749 & 
		0.093 & & 0.746  & 0.241 & & 0.852 & 0.247  \\
		\cmidrule{1-10}
		0.4 & 50 &  0.678 & 
		0.106  & & 0.779 & 0.240  & & 0.875  & 0.245  \\
		0.4 & 100 &  0.700 & 
		0.103 & & 0.772  & 0.241 & & 0.870 & 0.247  \\
		0.4 & 300 &  0.752 & 
		0.093 & & 0.738  & 0.242 & & 0.845 & 0.247  \\
		\cmidrule{1-10}
		0.6 & 50 &  0.706 & 
		0.103  & & 0.763  & 0.242  & & 0.861  & 0.246  \\
		0.6 & 100 &  0.720 & 
		0.100 & & 0.758  & 0.242  & & 0.857  & 0.247  \\
		0.6 & 300 &  0.753 & 
		0.093 & & 0.731  & 0.242  & & 0.838  & 0.247  \\
	\bottomrule
	\end{tabular}%
	\begin{tablenotes}[normal,flushleft]
		\begin{footnotesize}
	\item 
\!\!Columns represent: ``$\gamma$''/``$S$'': distance threshold/step threshold in shape guidance;
``JS. bond'': the JS divergence of bond length distributions of all the bond types between real molecules and generated molecules;
All the others are identical to those in Table~\ref{tbl:overall}. \par
		\par
		\end{footnotesize}
	\end{tablenotes}
\end{threeparttable}
\end{small}
  \vspace{-15pt}    
\end{table}


We conducted a parameter study to evaluate the impact of the distance threshold $\gamma$ (Eq.~\ref{eqn:guidance}) 
and the step threshold 
$S$ (Eq.~\ref{eqn:steps}) in the shape guidance.
%
%
Particularly, using the same trained \methoddiff module, we sampled molecules with 
different values of $\gamma$ and $S$ and present the results in 
Table~\ref{tbl:ablation_study:shape_guidance}. 
%
As shown in Table~\ref{tbl:ablation_study:shape_guidance}, the average shape similarities \avgshapesim \ and maximum shape similarities \maxshapesim \ 
consistently decrease as $\gamma$ and $S$ increase.
%
%
For example, when $S=50$, \avgshapesim \ and \maxshapesim \ decreases from 0.794 to 0.763 and 0.890 to 0.861, respectively, as $\gamma$ increases from $0.2$ to $0.6$.
Similarly, when $\gamma=0.2$, \avgshapesim \ and \maxshapesim \ decreases from 0.794 to 0.746 and 0.890 to 0.852, respectively, as $S$ increases from $50$ to $300$.
As presented in ``\method with Shape Guidance", larger $\gamma$ and $S$ indicate stronger shape guidance in \methodwithguide.
These results demonstrate that stronger shape guidance in \methodwithguide could effectively induce  
higher shape similarities between the given molecule and generated molecules.

It is also noticed that as shown in Table~\ref{tbl:ablation_study:shape_guidance}, incorporating shape guidance enables a trade-off between the quality of the generated molecules (\qed), 
and the shape similarities (\avgshapesim \ and \maxshapesim) between the given molecule and the generated ones.
For example, when $\gamma=0.2$, \qed increases from 0.630 to 0.749 and \avgshapesim \ decreases from 0.794 to 0.746 as $S$ increases from $50$ to $300$.
These results indicate the effects of $\gamma$ and $S$ in guiding molecule generation conditioned on given 
shapes. 
\subsubsection{Case Study}

\begin{figure}[!h]
		\vspace{-10pt}
		\centering
		\begin{minipage}{.1\linewidth}
			\vskip-55pt
		\end{minipage}
		\begin{subfigure}[b]{\linewidth}
			\centering
			\begin{subfigure}[b]{.4\linewidth}
				\centering
				\includegraphics[width=0.75\linewidth]{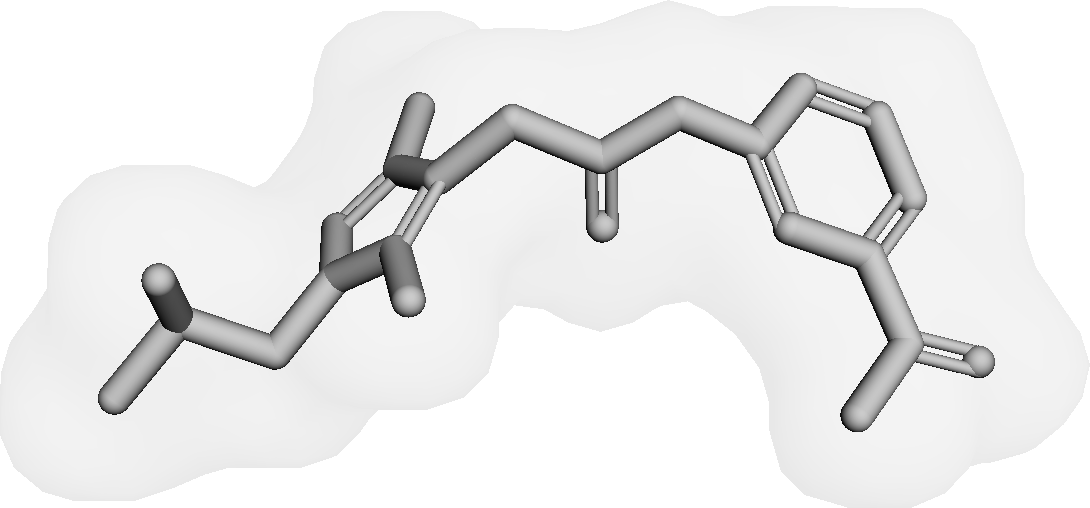}
			\end{subfigure}
			\begin{subfigure}[b]{.4\linewidth}
				\centering
				\includegraphics[width=.68\linewidth]{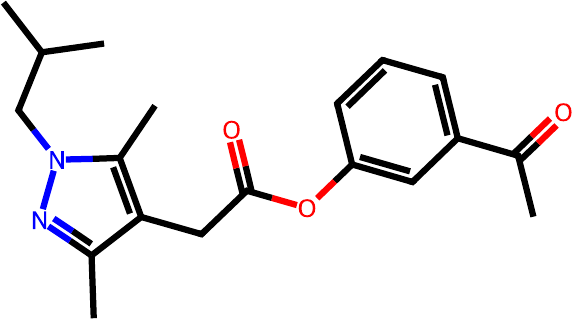}
			\end{subfigure}
			\captionsetup{justification=centering}
			\caption{condition molecule \molx, \qed = 0.462}
		\end{subfigure}
		\\
		\vspace{-10pt}
		\begin{minipage}{.1\linewidth}
			\vskip-55pt
		\end{minipage}
		\begin{subfigure}[b]{\linewidth}
			\centering
			\begin{subfigure}[b]{.3\linewidth}
				\centering
				\includegraphics[width=\linewidth]{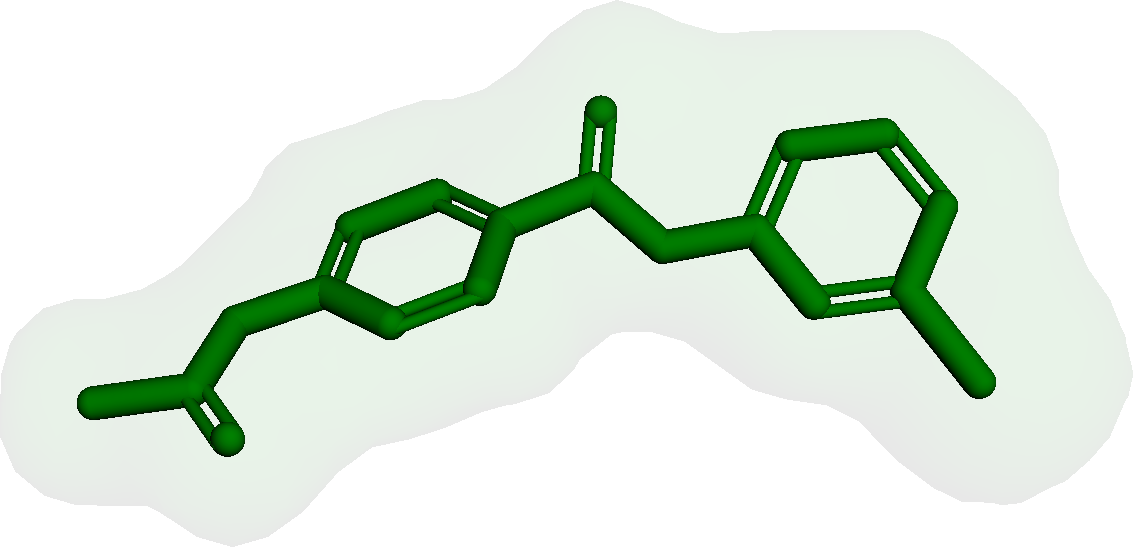}
			\end{subfigure}
			\begin{subfigure}[b]{.3\linewidth}
				\centering
				\includegraphics[width=\linewidth]{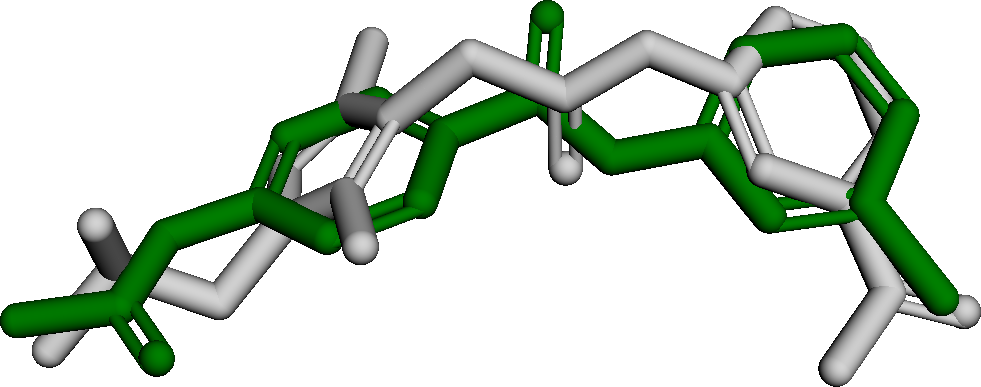}
			\end{subfigure}
			\begin{subfigure}[b]{.3\linewidth}
				\centering
				\includegraphics[width=\linewidth]{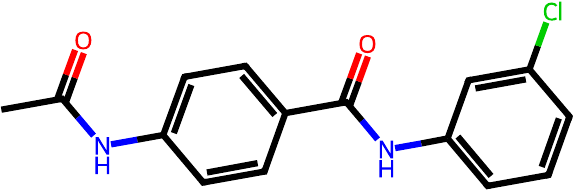}
			\end{subfigure}
			\captionsetup{justification=centering}
			\caption{\moly from \dataset: \shapesim = 0.759, \graphsim = 0.168, \qed = 0.907}
		\end{subfigure}
		\\
		\vspace{-10pt}
		\begin{minipage}{.1\linewidth}
			\vskip-55pt
		\end{minipage}
		\begin{subfigure}[b]{\linewidth}
			\centering
			\begin{subfigure}[b]{.3\linewidth}
				\centering
				\includegraphics[width=\linewidth]{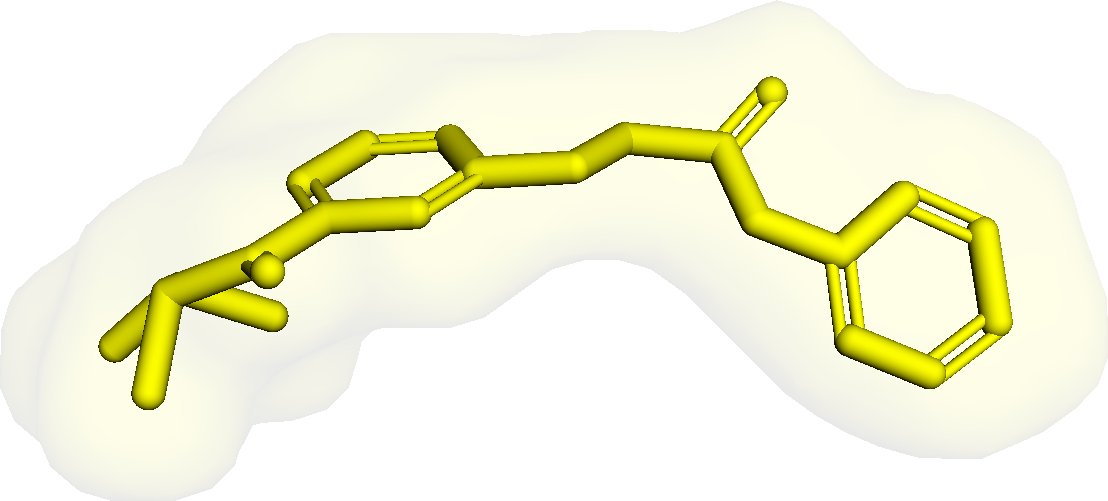}
			\end{subfigure}
			\begin{subfigure}[b]{.3\linewidth}
				\centering
				\includegraphics[width=\linewidth]{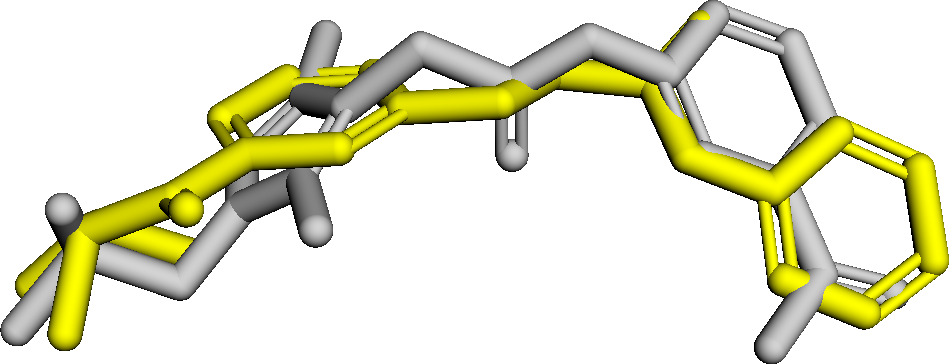}
			\end{subfigure}
			\begin{subfigure}[b]{.3\linewidth}
				\centering
				\includegraphics[width=\linewidth]{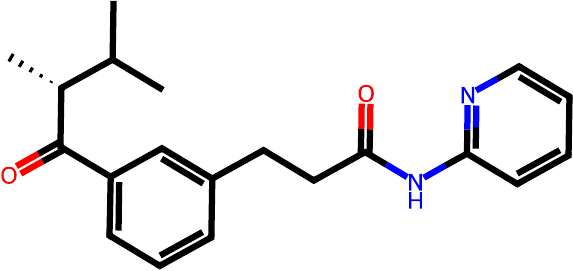}
			\end{subfigure}
			\captionsetup{justification=centering}
			\caption{\moly from \squid: \shapesim = 0.749, \graphsim = 0.243, \qed = 0.779}
		\end{subfigure}
		\\
		\begin{minipage}{.1\linewidth}
			\vskip-55pt
		\end{minipage}
		\begin{subfigure}[b]{\linewidth}
		\vspace{-10pt}		
			\centering
			\begin{subfigure}[b]{.3\linewidth}
				\centering
				\includegraphics[width=\linewidth]{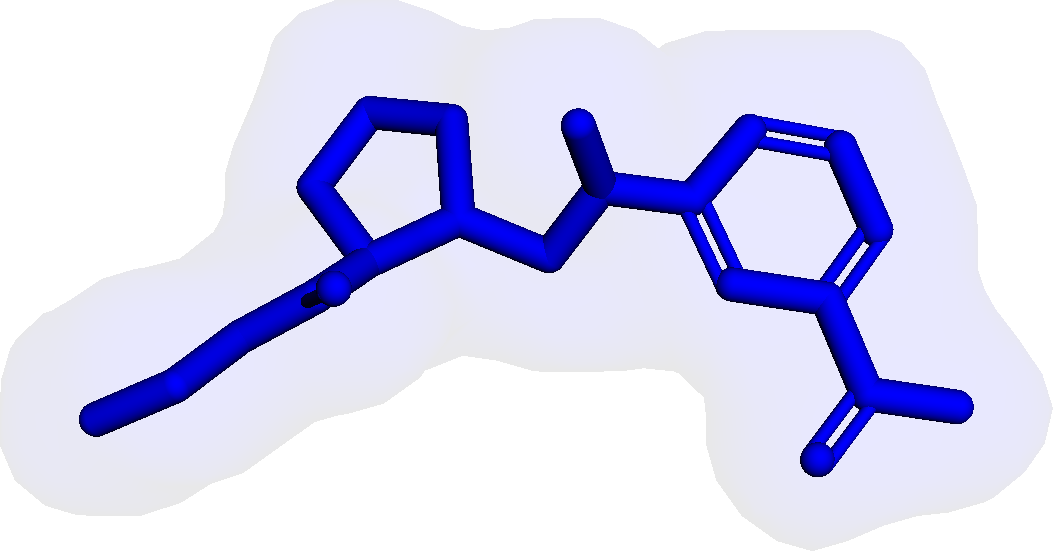}
			\end{subfigure}
			\begin{subfigure}[b]{.3\linewidth}
				\centering
				\includegraphics[width=\linewidth]{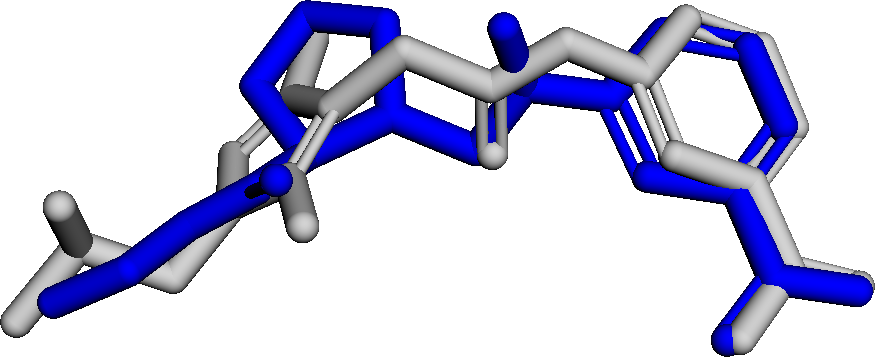}
			\end{subfigure}
			\begin{subfigure}[b]{.3\linewidth}
				\centering
				\includegraphics[width=\linewidth]{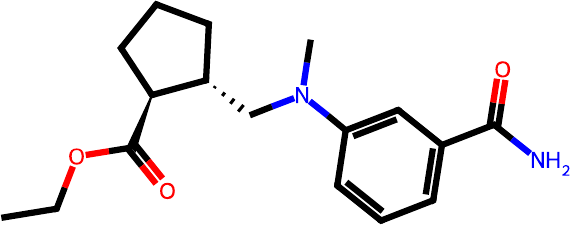}
			\end{subfigure}
			\captionsetup{justification=centering}
			\caption{\moly from \method: \shapesim = 0.835, \graphsim = 0.242, \qed = 0.818}
			\label{fig:ours}
		\end{subfigure}
		\vspace{-10pt}
		\caption{Generated 3D Molecules from Different Methods. 
		 Molecule 3D shapes are in shades; generated molecules are superpositioned with 
the condition molecule; and the molecular graphs of generated molecules are presented.}
		\label{fig:example1}
	\vspace{-20pt}
\end{figure}

Figure~\ref{fig:example1} presents three generated molecules from three methods given the same condition molecule. 
As shown in Figure~\ref{fig:example1}, the molecule generated by \method has higher shape similarity (0.835) with the condition molecule than those 
from the baseline methods (0.759 for \dataset and 0.749 for \squid).
Particularly, the molecule from \method has the surface shape (represented as blue shade in Figure~\ref{fig:ours}) most 
similar to that of the condition molecule. 
All three molecules have low graph similarities with the condition molecule and higher \qed scores
than the condition molecule.
This example shows the ability of \method to generate novel molecules that are more similar in 3D shape to condition molecules than those from baseline methods.
%

\section{Discussions and Conclusions}

In this paper, we develop a novel generative model \method, which generates 3D molecules conditioned on the 3D shape of given molecules.
\method utilizes a pre-trained equivariant shape encoder to generate equivariant embeddings for 3D shapes of given molecules.
Conditioned on the embeddings, \method learns an equivariant diffusion model to generate novel molecules.
%
%
%
To improve the shape similarities between the given molecule and the generated ones, we develop \methodwithguide, 
which incorporates shape guidance to push the generated atom positions to the shape of the given molecule.
We compare \method and \methodwithguide against state-of-the-art baseline methods.
Our experimental results demonstrate that \method and \methodwithguide could generate molecules with higher shape similarities, 
and competitive qualities compared to the baseline methods.
%
%
In future work, we will explore generating 3D molecules jointly conditioned on the shape 
and the electrostatic, considering that the electrostatic of molecules could also determine the binding activities of molecules.

%

\section*{Acknowledgements}

This project was made possible, in part, by support from 
{the National Science
Foundation grant nos. IIS-2133650 (X.N.), 
and} The Ohio State University President's Research Excellence
program (X.N.). Any opinions, findings and conclusions or recommendations
expressed in this paper are those of the authors and do not necessarily reflect the views of
the funding agency.

\bibliography{paper}

\clearpage



\section*{Supplementary material}

# Shape-conditioned 3D Molecule Generation via Equivariant Diffusion Models (Supplementary Materials)

Ziqi Chen[1], Bo Peng[1], Srinivasan Parthasarathy[1,2], Xia Ning[1,2,3*],

[1]Computer Science and Engineering, The Ohio Sate University, Columbus, OH 43210
[2]Translational Data Analytics Institute, The Ohio Sate University, Columbus, OH 43210
[3]Biomedical Informatics, The Ohio Sate University, Columbus, OH 43210
chen.8484@buckeyemail.osu.edu, peng.307@buckeyemail.osu.edu, srini@cse.ohio-state.edu, ning.104@osu.edu

## S1  Parameters for Reproducibility

We implemented both SE and DIFF using Python-3.7.16, PyTorch-1.11.0, PyTorch-scatter-2.0.9, Numpy-1.21.5, Scikit-learn-1.0.2. We trained the models using a Tesla V100 GPU with 32GB memory and a CPU with 80GB memory on Red Hat Enterprise 7.7. We released the code, data, and the trained model at Google Drive [1].

### S1.1  Parameters of SE

In SE, we tuned the dimension of all the hidden layers including VN-DGCNN layers (Eq. ), MLP layers (Eq. 1) and VN-In layer (Eq. 1), and the dimension $d_p$ of generated shape latent embeddings $\mathbf{H}^s$ with the grid-search algorithm in the parameter space presented in Table S1. We determined the optimal hyper-parameters according to the mean squared errors of the predictions of signed distances for 1,000 validation molecules that are selected as described in Section "Data" in the main manuscript. The optimal dimension of all the hidden layers is 128, and the optimal dimension $d_p$ of shape latent embedding $\mathbf{H}^s$ is 32. The optimal number of points $|\mathcal{P}|$ in the point cloud $\mathcal{P}$ is 512. We sampled 1,024 query points in $\mathcal{Z}$ for each molecule shape. We constructed graphs from point clouds, which are employed to learn $\mathbf{H}^s$ with VN-DGCNN layer (Eq. ), using the $k$-nearest neighbors based on Euclidean distance with $k=20$. We set the number of VN-DGCNN layers as 4. We set the number of MLP layers in the decoder (Eq. 1) as 4. We set the number of VN-In layers as 1.

We optimized the SE model via Adam (Kingma and Ba 2015) with its parameters (0.950, 0.999), learning rate 0.001, and batch size 16. We evaluated the validation loss every 2,000 training steps. We scheduled to decay the learning rate with a factor of 0.6 and a minimum learning rate of 1e-6 if the validation loss does not decrease in 5 consecutive evaluations. The optimal SE model has 27.3K learnable parameters. We trained the SE model with ∼172,600 training steps. The training took 80 hours with our GPUs. The trained SE model achieved the minimum validation loss at 162,000 steps.

---
*Contact author
[1]https://drive.google.com/drive/folders/146cpjuwenKGTd6Zh4sYBy-Wv6BMfGwe4?usp=sharing

Table S1: Hyper-Parameter Space for SE Optimization

| Hyper-parameters | Space |
|---|---|
| hidden layer dimension | {64, 128} |
| dimension $d_p$ of $\mathbf{H}^s$ | {32, 64} |
| # points in $\mathcal{P}$ | {512, 1024} |
| # query points in $\mathcal{Z}$ | 1024 |
| # nearest neighbors | 20 |
| # VN-DGCNN layers (Eq ) | 4 |
| # MLP layers in Eq 1 | 4 |

Table S2: Hyper-Parameter Space for DIFF Optimization

| Hyper-parameters | Space |
|---|---|
| hidden layer dimension | 128 |
| weight of atom type loss $\xi$ (Eq. 22) | 100 |
| threshold of step weight $\delta$ (Eq. 20) | 10 |
| # atom features $K$ | 15 |
| # EQ-GNN/INV-GNN layers | 8 |
| # heads $n_h$ in MHA$^x$/MHA$^v$ | 16 |
| # nearest neighbors $N$ (Eq. 16 and 18) | 8 |
| # diffusion steps $T$ | 1,000 |

### S1.2  Parameters of DIFF

Table S2 presents the parameters used to train DIFF. In DIFF, we set the dimension of all the hidden layers including MHA$^x$ and VN-Lin layer (Eq. 16), MHA$^h$ and VN-In layer (Eq. 18) and MLP layer (Eq. 19) as 128. We set the number of layers $L$ in EQ-GNN and INV-GNN as 8. Each layer is a multi-head attention layer (MHA$^x$ or MHA$^h$) with 16 heads. We set the number of VN-Lin and VN-In layer as 1, and the number of MLP layer as 2.

We constructed graphs from atoms in molecules, which are employed to predict atom coordinates and features (Eq. 16 and 18), using the $N$-nearest neighbors based on Euclidean distance with $N=8$. We used $K=15$ atom features in total, indicating the atom types and its aromaticity. These atom features include 10 non-aromatic atoms (i.e., "H", "C", "N", "O", "F", "P", "S", "Cl", "Br", "I"), and 5 aromatic atoms (i.e., "C", "N", "O", "P", "S"). We set the number of diffusion steps $T$ as 1,000. We set the weight $\xi$ of atom type loss (Eq. 22) as 100 to balance the values of atom type loss and atom coordinate loss. We set

the threshold $\delta$ (Eq. 20) as 10. The parameters $\beta_t^\mathtt{x}$ and $\beta_t^\mathtt{v}$ of variance scheduling in the forward diffusion process of DIFF are discussed in Supplementary Section S4.1. Following SQUID, we did not perform extensive hyperparameter tunning for DIFF given that the used hyperparameters have enabled good performance.

We optimized the DIFF model via Adam (Kingma and Ba 2015) with its parameters (0.950, 0.999), learning rate 0.001, and batch size 32. We evaluated the validation loss every 2,000 training steps. We scheduled to decay the learning rate with a factor of 0.6 and a minimum learning rate of 1e-5 if the validation loss does not decrease in 10 consecutive evaluations. The DIFF model has 2.7M learnable parameters. We trained the DIFF model with ∼900,000 training steps. The training took 60 hours with our GPUs. The trained DIFF achieved the minimum validation loss at 746,000 steps.

During inference, given a condition molecule, we determined the number of atoms for the molecule to be generated, according to the shape size of the condition molecule. We assumed that molecules with similar shape sizes should have similar numbers of atoms. Therefore, we identified the training molecules with similar shape sizes to the condition molecule. From these molecules, we built an atom number distribution and sampled the number of atoms in the molecule to be generated from it. Particularly, to calculate the shape size of each molecule, we first converted it into an atomic density grid of side length 22 Å with 0.5 Å resolution and then counted the non-empty voxels in this grid. For a condition molecule with shape size $n_s$, we selected training molecules with shape sizes in the range $[n_s - 200, n_s + 200]$ to build the atom number distribution. Following Adams and Coley (Adams and Coley 2023), we set the variance $\phi$ of atom-centered Gaussians as 0.049, which is used to build a set of points for shape guidance in Section "ShapeMol with Shape Guidance" in the main manuscript. The optimal distance threshold $\gamma$ is 0.2, and the optimal stop step $S$ for shape guidance is 300. With shape guidance, each time we updated the atom position (Eq. 23), we randomly sampled the weight $\sigma$ from $[0.2, 0.8]$. For each condition molecule, it took around 40 seconds on average to generate 50 molecule candidates with our GPUs.

## S2  Point Cloud Construction

In ShapeMol, we represented molecular surface shapes using point clouds ($\mathcal{P}$). $\mathcal{P}$ serves as input to ShapeMol-enc, from which we derive shape latent embeddings. To generate $\mathcal{P}$, we initially generated a molecular surface mesh using the algorithm from the Open Drug Discovery Toolkit (Wójcikowski, Zielenkiewicz, and Siedlecki 2015). Following this, we uniformly sampled points on the mesh surface with probability proportional to the face area, using the algorithm from PyTorch3D (Ravi et al. 2020). This point cloud $\mathcal{P}$ is then centralized by setting the center of its points to zero.

## S3  Query Point Sampling

As described in Section "Shape Decoder (SE-dec)", the signed distances of query points $z_q$ to molecule surface shape $\mathcal{P}$ are used to optimize SE. In this section, we present how to sample these points $z_q$ in 3D space. Particularly, we first determined the bounding box around the molecular surface shape, using the maximum and minimum $(x, y, z)$-axis coordinates for points in our point cloud $\mathcal{P}$, denoted as $(x_{\min}, y_{\min}, z_{\min})$ and $(x_{\max}, y_{\max}, z_{\max})$. We extended this box slightly by defining its corners as $(x_{\min} - 1, y_{\min} - 1, z_{\min} - 1)$ and $(x_{\max} + 1, y_{\max} + 1, z_{\max} + 1)$. For sampling $|\mathcal{Z}|$ query points, we wanted an even distribution of points inside and outside the molecule surface shape. When a bounding box is defined around the molecule surface shape, there could be a lot of empty spaces within the box that the molecule does not occupy due to its complex and irregular shape. This could lead to that fewer points within the molecule surface shape could be sampled within the box. Therefore, we started by randomly sampling $3k$ points within our bounding box to ensure that there are sufficient points within the surface. We then determined whether each point lies within the molecular surface, using an algorithm from Trimesh [2] based on the molecule surface mesh. If there are $n_w$ points found within the surface, we selected $n = \min(n_w, k/2)$ points from these points, and randomly choose the remaining $k - n$ points from those outside the surface. For each query point, we determined its signed distance to the molecule surface by its closest distance to points in $\mathcal{P}$ with a sign indicating whether it is inside the surface.

## S4  Forward Diffusion (DIFF-forward)

### S4.1  Variance Scheduling in DIFF-forward

Following Guan *et al.* (Guan et al. 2023), we used a sigmoid $\beta$ schedule for the variance schedule $\beta_t^\mathtt{x}$ of atom coordinates as below,

$$\beta_t^\mathtt{x} = \text{sigmoid}(w_1(2t/T - 1))(w_2 - w_3) + w_3 \quad (S1)$$

in which $w_i$($i$=1,2, or 3) are hyperparameters; $T$ is the maximum step. We set $w_1 = 6$, $w_2 = 1.e - 7$ and $w_3 = 0.01$. For atom types, we used a cosine $\beta$ schedule (Nichol and Dhariwal 2021) for $\beta_t^\mathtt{v}$ as below,

$$\begin{aligned}\bar{\alpha}_t^\mathtt{v} &= \frac{f(t)}{f(0)}, f(t) = \cos(\frac{t/T + s}{1 + s} \cdot \frac{\pi}{2})^2 \\ \beta_t^\mathtt{v} &= 1 - \alpha_t^\mathtt{v} = 1 - \frac{\bar{\alpha}_t^\mathtt{v}}{\bar{\alpha}_{t-1}^\mathtt{v}}\end{aligned} \quad (S2)$$

in which $s$ is a hyperparameter and set as 0.01.

As shown in Section "Forward Diffusion Process", the values of $\beta_t^\mathtt{x}$ and $\beta_t^\mathtt{v}$ should be sufficiently small to ensure the smoothness of forward diffusion process. In the meanwhile, their corresponding $\bar{\alpha}_t$ values should decrease from 1 to 0 over $t = [1, T]$. Figure S1 shows the values of $\beta_t$ and $\bar{\alpha}_t$ for atom coordinates and atom types with our hyperparameters. Please note that the value of $\beta_t^\mathtt{x}$ is less than 0.1 for 990 out of 1,000 steps. This guarantees the smoothness of the forward diffusion process.

---
[2]https://trimsh.org/

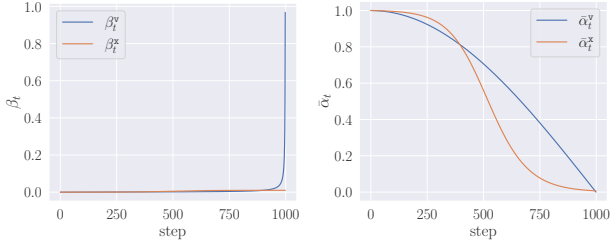

Figure S1: Schedule

## S4.2 Derivation of Forward Diffusion Process

In ShapeMol, a Gaussian noise and a categorical noise are added to continuous atom position and discrete atom features, respectively. Here, we briefly describe the derivation of posterior equations (i.e., Eq. 8, and 10) for atom positions and atom types in our work. We refer readers to Ho *et al.* (2020) and Kong *et al.* (2021) for a detailed description of diffusion process for continuous variables and Hoogeboom *et al.* (Hoogeboom et al. 2021) for the description of diffusion process for discrete variables.

For continuous atom positions, as shown in Kong *et al.* (2021), according to Bayes theorem, given $q(\mathbf{x}_t|\mathbf{x}_{t-1})$ defined in Eq. 3 and $q(\mathbf{x}_t|\mathbf{x}_0)$ defined in Eq. 5, the posterior $q(\mathbf{x}_{t-1}|\mathbf{x}_t, \mathbf{x}_0)$ is derived as below (superscript x is omitted for brevity),

$$
\begin{aligned}
q(\mathbf{x}_{t-1}|\mathbf{x}_t, \mathbf{x}_0) &= \frac{q(\mathbf{x}_t|\mathbf{x}_{t-1}, \mathbf{x}_0)q(\mathbf{x}_{t-1}|\mathbf{x}_0)}{q(\mathbf{x}_t|\mathbf{x}_0)} \\
&= \frac{\mathcal{N}(\mathbf{x}_t|\sqrt{1-\beta_t}\mathbf{x}_{t-1}, \beta_t\mathbf{I})\mathcal{N}(\mathbf{x}_{t-1}|\sqrt{\bar{\alpha}_{t-1}}\mathbf{x}_0, (1-\bar{\alpha}_{t-1})\mathbf{I})}{\mathcal{N}(\mathbf{x}_t|\sqrt{\bar{\alpha}_t}\mathbf{x}_0, (1-\bar{\alpha}_t)\mathbf{I})} \\
&= (2\pi\beta_t)^{-\frac{3}{2}}(2\pi(1-\bar{\alpha}_{t-1}))^{-\frac{3}{2}}(2\pi(1-\bar{\alpha}_t))^{\frac{3}{2}} \times \exp( \\
&\quad -\frac{\|\mathbf{x}_t - \sqrt{\alpha_t}\mathbf{x}_{t-1}\|^2}{2\beta_t} - \frac{\|\mathbf{x}_{t-1} - \sqrt{\bar{\alpha}_{t-1}}\mathbf{x}_0\|^2}{2(1-\bar{\alpha}_{t-1})} \\
&\quad + \frac{\|\mathbf{x}_t - \sqrt{\bar{\alpha}_t}\mathbf{x}_0\|^2}{2(1-\bar{\alpha}_t)}) \\
&= (2\pi\tilde{\beta}_t)^{-\frac{3}{2}}\exp(-\frac{1}{2\tilde{\beta}_t}\|\mathbf{x}_{t-1} - \frac{\sqrt{\bar{\alpha}_{t-1}}\beta_t}{1-\bar{\alpha}_t}\mathbf{x}_0 \\
&\quad - \frac{\sqrt{\alpha_t}(1-\bar{\alpha}_{t-1})}{1-\bar{\alpha}_t}\mathbf{x}_t\|^2)
\end{aligned}
$$
where $\tilde{\beta}_t = \frac{1-\bar{\alpha}_{t-1}}{1-\bar{\alpha}_t}\beta_t$.

(S3)

Therefore, the posterior of atom positions is derived as below,

$$q(\mathbf{x}_{t-1}|\mathbf{x}_t, \mathbf{x}_0) = \mathcal{N}(\mathbf{x}_{t-1}|\frac{\sqrt{\bar{\alpha}_{t-1}}\beta_t}{1-\bar{\alpha}_t}\mathbf{x}_0 + \frac{\sqrt{\alpha_t}(1-\bar{\alpha}_{t-1})}{1-\bar{\alpha}_t}\mathbf{x}_t, \tilde{\beta}_t\mathbf{I}). \quad (S4)$$

For discrete atom features, as shown in Hoogeboom *et al.* (Hoogeboom et al. 2021) and Guan *et al.* (Guan et al. 2023), according to Bayes theorem, the posterior $q(\mathbf{v}_{t-1}|\mathbf{v}_t, \mathbf{v}_0)$ is derived as below (supperscript v is omitted for brevity),

$$q(\mathbf{v}_{t-1}|\mathbf{v}_t, \mathbf{v}_0) = \frac{q(\mathbf{v}_t|\mathbf{v}_{t-1}, \mathbf{v}_0)q(\mathbf{v}_{t-1}|\mathbf{v}_0)}{\sum_{\mathbf{v}_{t-1}}q(\mathbf{v}_t|\mathbf{v}_{t-1}, \mathbf{v}_0)q(\mathbf{v}_{t-1}|\mathbf{v}_0)} \quad (S5)$$

For $q(\mathbf{v}_t|\mathbf{v}_{t-1}, \mathbf{v}_0)$, we have

$$
\begin{aligned}
q(\mathbf{v}_t|\mathbf{v}_{t-1}, \mathbf{v}_0) &= \mathcal{C}(\mathbf{v}_t|(1-\beta_t)\mathbf{v}_{t-1} + \beta_t/K) \\
&= \begin{cases} 1-\beta_t+\beta_t/K, & \text{when } \mathbf{v}_t = \mathbf{v}_{t-1}, \\ \beta_t/K, & \text{when } \mathbf{v}_t \neq \mathbf{v}_{t-1}, \end{cases} \\
&= \mathcal{C}(\mathbf{v}_{t-1}|(1-\beta_t)\mathbf{v}_t + \beta_t/K).
\end{aligned}
$$
(S6)

Therefore, we have

$$
\begin{aligned}
&q(\mathbf{v}_t|\mathbf{v}_{t-1}, \mathbf{v}_0)q(\mathbf{v}_{t-1}|\mathbf{v}_0) \\
&= \mathcal{C}(\mathbf{v}_{t-1}|(1-\beta_t)\mathbf{v}_t + \beta_t\frac{1}{K})\mathcal{C}(\mathbf{v}_{t-1}|\bar{\alpha}_{t-1}\mathbf{v}_0 + (1-\bar{\alpha}_{t-1})\frac{1}{K}) \\
&= [\alpha_t\mathbf{v}_t + \frac{1-\alpha_t}{K}] \odot [\bar{\alpha}_{t-1}\mathbf{v}_0 + \frac{1-\bar{\alpha}_{t-1}}{K}].
\end{aligned}
$$
(S7)

Therefore, with $q(\mathbf{v}_t|\mathbf{v}_{t-1}, \mathbf{v}_0)q(\mathbf{v}_{t-1}|\mathbf{v}_0) = \tilde{\mathbf{c}}$, the posterior is as below,

$$q(\mathbf{v}_{t-1}|\mathbf{v}_t, \mathbf{v}_0) = \mathcal{C}(\mathbf{v}_{t-1}|\mathbf{c}(\mathbf{v}_t, \mathbf{v}_0)) = \frac{\tilde{\mathbf{c}}}{\sum_k^K \tilde{c}_k}. \quad (S8)$$

## S5 Backward Generative Process (DIFF-backward)

### S5.1 Proof of Equivariance in Atom Coordinate Prediction

In this section, we prove that our design in Eq. 16 in the main manuscript updates atom coordinates in an equivariant way. A function $f(\mathbf{x})$ is equivariant with respect to rotation, if given any rotation matrix $\mathbf{R} \in \mathbb{R}^{3\times 3}$, it satisfies,

$$f(\mathbf{Rx}) = \mathbf{R}f(\mathbf{x}). \quad (S9)$$

In Eq. 16, the embedding $\Delta\mathbf{x}_i^{l+1} \in \mathbb{R}^{n_h \times 3}$, which aggregates the neighborhood information of $a_i$, is updated by atom coordinates $\mathbf{x}_i^l$ from previous layer and embeddings from multi-head attention layer $\text{MHA}^\mathbf{x}(\cdot)$ as below,

$$\Delta\mathbf{x}_i^{l+1} = \sum_{j\in N(a_i), i\neq j}(\mathbf{x}_i^l - \mathbf{x}_j^l)\text{MHA}^\mathbf{x}(d_{ij}^l, \mathbf{h}_i^{l+1}, \mathbf{h}_j^{l+1}, \text{VN-In}(\mathbf{H}^\mathbf{s})),$$

$$\mathbf{x}_i^{l+1} = \mathbf{x}_i^l + \text{Mean}(\Delta\mathbf{x}_i^{l+1}) + \text{VN-Lin}(\mathbf{x}_i^l, \Delta\mathbf{x}_i^{l+1}, \mathbf{H}^\mathbf{s}).$$

The embeddings from $\text{MHA}^\mathbf{x}(\cdot)$ and atom feature embeddings $\mathbf{h}_i^l$ will be proved to be invariant in Supplementary Section S5.2 later. Therefore, $\text{MHA}^\mathbf{x}(\cdot)$ and $\mathbf{h}_i^{l+1}$ remain the same with rotation $\mathbf{R}$. Then, we can prove that $\Delta\mathbf{x}_i^{l+1}$ is updated in an equivariant way as below,

$$
\begin{aligned}
&\sum_{j\in N(a_i), i\neq j}(\mathbf{Rx}_i^l - \mathbf{Rx}_j^l)\text{MHA}^\mathbf{x}(d_{ij}^l, \mathbf{h}_i^{l+1}, \mathbf{h}_j^{l+1}, \text{VN-In}(\mathbf{H}^\mathbf{s})) \\
&= \mathbf{R}\sum_{j\in N(a_i), i\neq j}(\mathbf{x}_i^l - \mathbf{x}_j^l)\text{MHA}^\mathbf{x}(d_{ij}^l, \mathbf{h}_i^{l+1}, \mathbf{h}_j^{l+1}, \text{VN-In}(\mathbf{H}^\mathbf{s})) \\
&= \mathbf{R}\Delta\mathbf{x}_i^{l+1}.
\end{aligned}
$$

Given that mean pooling is a linear operation that respects rotation equivariance, $\text{Mean}(\Delta \mathbf{x}_i^{l+1})$ is equivariant to rotations. The vector-neuron-based linear layer VN-Lin($\cdot$) with leaky-ReLu activation function also produces rotational equivariant embeddings according to Deng et al. (Deng et al. 2021). Therefore, we prove that atom coordinates $\mathbf{x}_i^{l+1}$ are updated in an equivariant way as below,

$$\begin{aligned}
&\mathbf{R}\mathbf{x}_i^l + \text{Mean}(\mathbf{R}\Delta\mathbf{x}_i^{l+1}) + \text{VN-Lin}(\mathbf{R}\mathbf{x}_i^l, \mathbf{R}\Delta\mathbf{x}_i^{l+1}, \mathbf{R}\mathbf{H}^s)) \\
&= \mathbf{R}\mathbf{x}_i^l + \mathbf{R}\text{Mean}(\Delta\mathbf{x}_i^{l+1}) + \mathbf{R}\text{VN-Lin}(\mathbf{x}_i^l, \Delta\mathbf{x}_i^{l+1}, \mathbf{H}^s) \\
&= \mathbf{R}(\mathbf{x}_i^l + \text{Mean}(\Delta\mathbf{x}_i^{l+1}) + \text{VN-Lin}(\mathbf{x}_i^l, \Delta\mathbf{x}_i^{l+1}, \mathbf{H}^s)) \\
&= \mathbf{R}\mathbf{x}_i^{l+1}
\end{aligned} \quad (S10)$$

## S5.2 Proof of Invariance in Atom Feature Prediction

In this section, we prove that our design in Eq. 18 in the main manuscript updates atom features in an invariant way. A function $f(\mathbf{x})$ is invariant with respect to rotation, if it satisfies,

$$f(\mathbf{R}\mathbf{x}) = f(\mathbf{x}). \quad (S11)$$

Recall that in Eq. 18, atom feature embeddings $\mathbf{h}_i^{l+1}$ is updated according to atomic distances $d_{ij}^l$, embeddings $\mathbf{h}_i$ from previous layer and invariant shape embeddings VN-In($\mathbf{H}^s$) from the vector-neuron based invariant layer adapted by (Chen et al. 2022) as follows,

$$\mathbf{h}_i^{l+1} = \mathbf{h}_i^l + \sum_{j \in N(a_i), i \neq j} \text{MHA}^{\mathbf{h}}(d_{ij}^l, \mathbf{h}_i^l, \mathbf{h}_j^l, \text{VN-In}(\mathbf{H}^s)), \mathbf{h}_i^0 = \mathbf{v}_i.$$

Atomic distances are invariant to transformations. Atom feature embeddings are initialized with atom feature vectors, which are also invariant to transformations. The VN-In(.) layer maps the equivariant shape embedding $\mathbf{H}^s \in \mathbb{R}^{d_p \times 3}$ into the invariant shape embedding $\mathbf{h}^s \in \mathbb{R}^{d_p}$ as below,

$$\mathbf{h}_c^s = \text{MLP}(\langle \mathbf{H}_c^s, \frac{\bar{\mathbf{H}}^s}{\|\bar{\mathbf{H}}^s\|}\rangle).$$

where $\bar{\mathbf{H}}^s = \frac{1}{d_p}\sum_{c=1}^{d_p} \mathbf{H}_c^s$ and $\mathbf{H}_c^s$ is the $c$-th channel of $\mathbf{H}^s$; $\mathbf{h}_c^s$ denotes the $c$-th channel of $\mathbf{h}^s$; $\langle \cdot \rangle$ is the inner-product operation which the results will not be changed by any rotation vector. Given that all the variables in the right-hand side of Eq. 18 are invariant to rotations, we prove that the atom feature embeddings are updated in an invariant way. Therefore, the invariant atom feature embeddings lead to invariant atom feature predictions in Eq. 19. Similarly, we can prove that the embeddings from MHA$^{\mathbf{x}}$ layer (Eq. 16) are invariant to rotations.

## S6 ShapeMol Loss Function Derivation

In this section, we demonstrate that a step weight $w_t^x$ based on the signal-to-noise ratio $\lambda_t$ should be included into the atom position loss (Eq. 20). In the diffusion process for continuous variables, the optimization problem is defined as below (Ho, Jain, and Abbeel 2020),

$$\begin{aligned}
&\arg\min_{\Theta} KL(q(\mathbf{x}_{t-1}|\mathbf{x}_t, \mathbf{x}_0) | p_{\Theta}(\mathbf{x}_{t-1}|\mathbf{x}_t, \tilde{\mathbf{x}}_{0,t})) \\
&= \arg\min_{\Theta} \frac{\bar{\alpha}_{t-1}(1-\alpha_t)}{2(1-\bar{\alpha}_{t-1})(1-\bar{\alpha}_t)}\|\tilde{\mathbf{x}}_{0,t} - \mathbf{x}_0\|^2 \\
&= \arg\min_{\Theta} \frac{1-\alpha_t}{2(1-\bar{\alpha}_{t-1})\alpha_t}\|\tilde{\boldsymbol{\epsilon}}_{0,t} - \boldsymbol{\epsilon}_0\|^2,
\end{aligned}$$

where $\boldsymbol{\epsilon}_0 = \frac{\mathbf{x}_t - \sqrt{\bar{\alpha}_t}\mathbf{x}_0}{\sqrt{1-\bar{\alpha}_t}}$ is the ground-truth noise variable sampled from $\mathcal{N}(\mathbf{0}, \mathbf{1})$ and is used to sample $\mathbf{x}_t$ from $\mathcal{N}(\mathbf{x}_t|\sqrt{\bar{\alpha}_t}\mathbf{x}_0, (1-\bar{\alpha}_t)\mathbf{I})$ in Eq. 6; $\tilde{\boldsymbol{\epsilon}}_0 = \frac{\mathbf{x}_t - \sqrt{\bar{\alpha}_t}\tilde{\mathbf{x}}_{0,t}}{\sqrt{1-\bar{\alpha}_t}}$ is the predicted noise variable.

Ho et al. (Ho, Jain, and Abbeel 2020) further simplified the above objective as below and demonstrated that the simplified one can achieve better performance:

$$\begin{aligned}
&\arg\min_{\Theta} \|\tilde{\boldsymbol{\epsilon}}_{0,t} - \boldsymbol{\epsilon}_0\|^2 \\
&= \arg\min_{\Theta} \frac{\bar{\alpha}_t}{1-\bar{\alpha}_t}\|\tilde{\mathbf{x}}_{0,t} - \mathbf{x}_0\|^2,
\end{aligned} \quad (S12)$$

where $\lambda_t = \frac{\bar{\alpha}_t}{1-\bar{\alpha}_t}$ is the signal-to-noise ratio. While previous work (Guan et al. 2023) applies uniform step weights across different steps, we demonstrate that a step weight should be included into the atom position loss according to Eq. S12. However, the value of $\lambda_t$ could be very large when $\bar{\alpha}_t$ is close to 1 as $t$ approaches 1. Therefore, we clip the value of $\lambda_t$ with threshold $\delta$ in Eq. 20.

## S7 Evaluation Metrics

We calculated the shape similarity $\text{Sim}_s(\mathbf{s}_x, \mathbf{s}_y)$ via the overlap volumes between two aligned molecules as in the literature (Adams and Coley 2023). Each molecule candidate $M_y$ for evaluation is aligned with the condition molecule $M_x$ by the ShaEP tool (Vainio, Puranen, and Johnson 2009). For the molecular graph similarity $\text{Sim}_g(M_x, M_y)$, we used the Tanimoto similarity over Morgan fingerprints between $M_x$ and $M_y$ calculated by RDKit (Landrum et al. 2023).



\end{document}